%% file: detikzify.tex
\pdfoutput=1
\PassOptionsToPackage{cachedir=minted,frozencache}{minted}
\documentclass{article} 
\usepackage[final]{styles/neurips_2024}
\usepackage{graphicx}
\usepackage{booktabs,wrapfig,widetable}
\usepackage[most,minted]{tcolorbox}
\usepackage{forest}
\usepackage{anyfontsize} 
\usepackage{hyperref}
\usepackage{styles/detikzify} 
\usepackage[autopunct]{csquotes}
\usetikzlibrary{calc,arrows.meta,decorations.pathmorphing}

\graphicspath{{graphics}{graphics/examples}{graphics/openmoji}}
\makeatletter\let\input@path\Ginput@path\makeatother

\hypersetup{%
  colorlinks=true,
  citecolor=darkblue,
  linkcolor=darkblue,
  urlcolor=darkblue
}

\title{\projectname: Synthesizing Graphics Programs\\for Scientific Figures and
Sketches with \tikzname}

\author{%
  Jonas Belouadi\footnotemark
  \And
  Simone Paolo Ponzetto\footnotemark
  \And
  Steffen Eger\footnotemark
  \AND\\[\dimexpr-24pt-1pt-\dp\strutbox]
  {\normalfont\href{https://nl2g.github.io}{\normalcolor Natural Language Learning Group}\dblfootnotemark[,]{1}{3}
  \href{https://www.uni-mannheim.de/dws}{\normalcolor Data and Web Science Group}\footnotemark[2]}\\
  University of Mannheim\dblfootnotemark[,]{1}{2} University of Technology Nuremberg\footnotemark[3]\\
  \mails{jonas.belouadi,ponzetto}{uni-mannheim.de}, \mail{steffen.eger@utn.de}%
  \setcounter{footnote}{0}%
}

\begin{document}

\maketitle

\begin{abstract}
   Creating high-quality scientific figures can be time-consuming and
   challenging, even though sketching ideas on paper is relatively easy.
   Furthermore, recreating existing figures that are not stored in formats
   preserving semantic information is equally complex.
   To tackle this problem, we introduce \projectname, a novel multimodal
   language model that automatically synthesizes scientific figures as
   semantics-preserving \tikzname graphics programs based on sketches and
   existing figures. To achieve this, we create three new datasets:
   \datikz[v2], the largest \tikzname dataset to date, containing over 360k
   human-created \tikzname graphics; \sketchfig, a dataset that pairs
   hand-drawn sketches with their corresponding scientific figures; and
   \metafig, a collection of diverse scientific figures and associated
   metadata.
  We train \projectname on \metafig and \datikz[v2], along with synthetically
  generated sketches learned from \sketchfig. We also introduce an \mcts-based
  inference algorithm that enables \projectname to iteratively refine its
  outputs without the need for additional training.
  Through both automatic and human evaluation, we demonstrate that \projectname
  outperforms commercial \claude and \gpt in synthesizing \tikzname programs,
  with the \mcts algorithm effectively boosting its performance.
  We make our code, models, and datasets publicly
  available.\footnote{\anonymize{\url{https://github.com/potamides/DeTikZify}}}
\end{abstract}
	
\input{sections/introduction}
\input{sections/related}
\input{sections/data}
\input{sections/models}
\input{sections/mcts}
\input{sections/experiments}
\input{sections/analysis}
\input{sections/conclusion}
\input{sections/limitations}
\anonymize[]{\input{sections/acknowledgements}}

\bibliography{detikzify}
\bibliographystyle{styles/acl_natbib}

\clearpage\appendix
\input{sections/appendix}
\makeatletter\unless\if@preprint
\clearpage
\fi\makeatother

\end{document}

%% file: sections/introduction.tex
\section{Introduction}\label{sec:introduction} 
Creating high-quality scientific figures is similar to typesetting scientific
documents in many ways. When it comes to typesetting, markup languages like
\latex enjoy widespread popularity, as exemplified by major machine learning
conferences that either mandate or strongly encourage \latex-formatted
submissions.\footnote{\UrlFont{https://\{%
  \href{https://neurips.cc/Conferences/2024/CallForPapers}{\nolinkurl{neurips.cc}},%
  \href{https://icml.cc/Conferences/2024/CallForPapers}{\nolinkurl{icml.cc}},%
  \href{https://iclr.cc/Conferences/2024/CallForPapers}{\nolinkurl{iclr.cc}}%
  \}/Conferences/2024/CallForPapers%
}}
The advantages of using such languages go beyond producing high-quality
outputs; documents expressed as high-level, semantics-preserving programs
enhance accessibility, serve archival purposes, and remain easily editable and
human-readable~(facilitating language modeling applications;
\citealp{moosavi2021scigen,lu2023scitab}). Consequently, efforts have been made
to recover this type of information from outputs stored in lower-level vector
graphics formats like \pdf or \svg, or raster graphics
formats~\citep{desai2021table,blecher2024nougat}. At the other
end of the spectrum, the versatility of \latex comes with a steep learning
curve, and typesetting can often be challenging for end users. In response,
researchers have been working on assisting authors with certain aspects of the
problem, such as typesetting math based on hand-drawn
sketches~\citep{kirsch2010detexify,wu2020math}.

Just like documents, scientific figures can also be created using markup
languages. A popular example is the \tikzname graphics
language~\citep{tantau2023tikz}, which can be integrated into \latex documents,
providing comparable benefits and encountering similar challenges. However,
unlike \latex, the prospects of \tikzname in research contexts remain largely
unexplored. Although the promise of simplifying editing and enabling
applications in visual
understanding~\citep{masry2022chartqa,huang2023summaries} is evident, there are
currently no viable solutions for recovering graphics programs from compiled
figures. Moreover, there is a lack of tools that assist in creating graphics
programs, e.g., based on hand-drawn sketches, despite the clear demand for such
approaches on the
\texse*~(\texse),\footnote{\url{https://tex.stackexchange.com}} where nearly
10\% of all questions revolve around \tikzname, making it the most frequently
discussed topic on the site. Addressing this gap could greatly improve the
accessibility of existing figures and support researchers at all levels of
programming proficiency when creating new ones, fostering diversity and
inclusion.
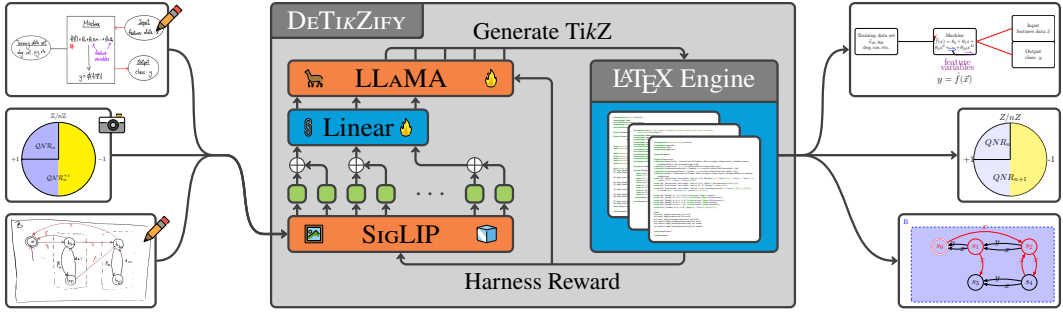
\begin{figure}
  \centering
  \input{architecture.tex}
  \caption{Overview of the \projectname architecture: A multimodal language
  model converts sketches or figures into \tikzname programs, which are
  compiled by a \latex engine. This provides a reward signal to the model via
  \mcts, allowing it to iteratively refine the output until satisfactory
  results are achieved.}%
  \label{fig:detikzify}
\end{figure}
In response, we introduce \projectname, a multimodal language model that
automatically synthesizes \tikzname programs for scientific figures and
sketches (cf.\ \figref{fig:detikzify}). Our key contributions are as follows:
\begin{enumerate}
  \item[(i)] As part of \projectname, we introduce (a) \datikz[v2], a large
    \tikzname dataset with over 360k human-created \tikzname graphics;
    (b) \sketchfig, a dataset of human-created sketches with paired
    scientific figures; and (c) \metafig, a large meta-dataset of scientific
    figures and associated texts.
  \item[(ii)] We train \projectname on \metafig and \datikz[v2], augmented with
    synthetic sketches that mimic \sketchfig. We demonstrate that \projectname
    can effectively synthesize \tikzname programs for both existing scientific
    figures and sketches, outperforming the commercial \llm*{}s~(\llm{}s) \gpt
    and \claude~\citep{openai2023gptv,anthropic2024claude}.
\item[(iii)] We also present an inference algorithm based on \mcts*~(\mcts)
    that is tailored to graphics programs and allows \projectname to
    iteratively refine \emph{its own outputs} for a given computational budget,
    further improving performance without additional training.
\end{enumerate}

%% file: graphics/architecture.tex
\begingroup%

\newlength\tcoverlap%
\setlength{\tcoverlap}{0.1pt}%

\newsavebox\pencil%
\newsavebox\camera%
\sbox{\pencil}{\includegraphics[width=3.5mm]{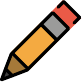}}%
\sbox{\camera}{\reflectbox{\includegraphics[width=3.5mm]{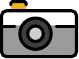}}}%

\tcbsetforeverylayer{%
  enhanced,%
  size=small,%
  halign=center,%
  valign=center,%
  center title,%
}%

\newtcbox{\model}[3][]{%
  capture=minipage,%
  overlay={%
    \node[anchor=west,outer sep=1mm+1mm+0.3mm] at (frame.west) {%
      \IfStrEq{#2}{}{}{\includegraphics[height=\tcbtextheight]{#2}}%
    };%
    \node[anchor=east,outer sep=1mm+1mm+0.3mm] at (frame.east) {%
      \IfStrEq{#3}{}{}{\includegraphics[height=\tcbtextheight]{#3}}%
    };%
  },%
  colback=TUDa-8a,%
  #1%
}%

\newtcbox{\patch}[1][]{%
  left=0pt, right=0pt, top=0pt, bottom=0pt,%
  colback=TUDa-4a,%
  #1%
}%

\newlength\tcbprevtextheight%
\newtcbinputlisting{\tikzlisting}[3][]{
  colback=white,%
  fontupper=\tcbfontsize{#2},%
  left=0pt,%
  right=0pt,%
  top=0pt,%
  bottom=0pt,%
  boxsep=0.5mm,%
  listing only,%
  minted language=latex,%
  minted options={%
    breaklines=true,%
    breaksymbol={},%
    breakaftersymbolpre={},%
    breakafter={,},%
    breakindentnchars=4%
  },%
  listing file={#3},%
  after={\global\tcbprevtextheight=\tcbtextheight},%
  #1%
}%

\newtcbox{\scifig}[2][]{%
  left=0pt,%
  right=0pt,%
  top=0pt,%
  bottom=0pt,%
  boxsep=0.5mm,%
  colback=white,%
  finish={%
    \node[anchor=center] at ($(frame.north east)!1/3!(frame.east)$) {#2};%
  },%
  #1%
}%

\newtcolorbox{component}[2][]{%
  colback=TUDa-2a,%
  title filled,%
  title={#2},%
  #1%
}%

\newtcolorbox{system}[2][]{%
  colback=TUDa-0a,%
  capture=hbox,%
  top=0.3mm-\tcboxedtitleheight,%
  attach boxed title to top left={%
    xshift=0.3mm-\tcoverlap,
    yshift*=-\tcboxedtitleheight-0.3mm+\tcoverlap%
  },%
  boxed title style={sharp corners=uphill,%
    size=small,%
    no borderline,%
    leftrule=\tcoverlap,%
    toprule=\tcoverlap%
  },%
  title={#2},%
  #1%
}%

\tikzset{%
  XOR/.style={%
    line width=0.15mm,%
    inner sep=0.2em,%
    fill=white,%
    draw=tcbcolframe,%
    circle,%
    append after command={%
      (\tikzlastnode.north) edge[line width=0.15mm,draw=tcbcolframe] (\tikzlastnode.south)%
      (\tikzlastnode.east) edge[line width=0.15mm,draw=tcbcolframe] (\tikzlastnode.west)%
    }%
  },%
  line/.style={%
    ->,%
    >={Stealth[round,width=1.5mm,length=1mm]},%
    rounded corners=0.75mm,%
    draw=tcbcolframe,%
    shorten <=-\tcoverlap%
  },%
  every node/.style={%
    inner sep=0pt,%
    outer sep=0pt%
  }%
}%

\newdimen\XCoord%
\newdimen\YCoord%
\newcommand*{\ExtractCoordinate}[1]{\path[] (#1); \pgfgetlastxy{\XCoord}{\YCoord};}%


\begin{system}[remember as=detikzify]{\projectname}%
\begin{tikzpicture}[remember picture,line width=.3mm]%
  \node at (0, 0) (vision) {\model[width=3cm]{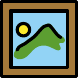}{ice.pdf}{\siglip}};

  \node[anchor=south west] at ($(vision.north west)+(0, 0.15)$) (patch1) {\patch{}};
  \node[anchor=east] at (vision.north east |- patch1) (patch6) {\patch{}};
  \node[anchor=center] at ($(patch1)!1/7!(patch6)$) (patch2) {\patch{}};
  \node[anchor=center] at ($(patch1)!2/7!(patch6)$) (patch3) {\patch{}};
  \node[anchor=center] at ($(patch1)!3/7!(patch6)$) (patch4) {\patch{}};
  \node[anchor=center] at ($(patch1)!6/7!(patch6)$) (patch5) {\patch{}};
  \node[anchor=center] at ($(patch4)!.5!(patch5)$) {\textcolor{tcbcolframe}{$\ldots$}};

  \foreach \i in {1,...,6} {
    \draw[line] (vision.north -| patch\i) to (patch\i.south);
  }

  \foreach \i [evaluate=\i as \j using int(\i+1)] in {1,3,5} {
    \node[anchor=south,XOR] at ($(patch\i.north)+(0, 0.15)$) (concat\i) {};
    \draw[line] (patch\i.north) to (concat\i.south);
    \draw[line] (patch\j.north) |- (concat\i.east);
  }

  \node[anchor=south west] at ($(vision.west |- concat1.north)+(0, 0.15)$) (linear) {%
    \model[width=1.875cm,colback=TUDa-2a]{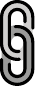}{fire.pdf}{Linear}%
  };

  \coordinate (linear1) at (linear.south -| concat1.north);
  \coordinate (linear3) at (linear.south -| concat3.north);
  \coordinate (linear5) at ($(linear.south -| patch4)+(patch4)-(patch3)$);

  \draw[line] (concat1.north) to (linear1);
  \draw[line] (concat3.north) to (linear3);
  \draw[line] (concat5.west) -| (linear5);

  \node[anchor=south west] at ($(linear.north west)+(0, 0.15)$) (text) {%
    \model[width=3cm]{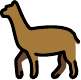}{fire.pdf}{\llama}%
  };

  \foreach \i in {1,3,5} {
    \draw[line] (linear.north -| linear\i) to (text.south -| linear\i);
  }


  \ExtractCoordinate{$(text.north)-(vision.south)$}
  \node[anchor=north west] at ($(text.north east)+(1, 0)$) (latex) {%
    \begin{component}[width=2.5cm, height=\YCoord,valign=top, halign=left]{\latex Engine}%
      \makebox(0,0)[lt]{%
        \begin{minipage}[t][\tcbtextheight][t]{\tcbtextwidth}%
          \tikzlisting[width=1.5cm]{0.075}{graphics/examples/output3_tex.tex}%
        \end{minipage}%
      }%
      \makebox(0,0)[lt]{%
        \begin{minipage}[t][\tcbtextheight][c]{\tcbtextwidth}%
          \centerline{%
            \tikzlisting[width=1.5cm,valign=top,text height=\tcbprevtextheight]{0.075}{%
              graphics/examples/output2_tex.tex}%
            }%
        \end{minipage}%
      }%
      \makebox(0,0)[lt]{%
        \begin{minipage}[t][\tcbtextheight][b]{\tcbtextwidth}%
          \rightline{%
            \tikzlisting[width=1.5cm,valign=top,text height=\tcbprevtextheight]{0.075}{%
              graphics/examples/output1_tex.tex}%
            }%
        \end{minipage}%
      }%
    \end{component}%
  };

  \coordinate (tokenpos) at (text.north -| {$(patch1)!3/7!(patch6)$});
  \draw[line] (tokenpos) to ($(tokenpos)+(0,0.15)$) to ($(text.north)+(0,0.15)$) --
    node [inner sep=.3333em, above] {\small Generate \tikzname} ($(latex.north)+(0,0.15)$) to (latex.north);
  \foreach \i in {4,...,7} {
    \coordinate (tokenpos) at (text.north -| {$(patch1)!\i/7!(patch6)$});
    \draw[line,-] (tokenpos) to ($(tokenpos)+(0,0.15)$);
  }

  \draw[line] (latex.south) to ($(latex.south)-(0,0.15)$) --
    node [inner sep=.3333em, below] {\small Harness Reward} ($(vision.south)-(0,0.15)$) to (vision.south);
  \draw[line] ({$(latex.south)-(0,0.15)$} -| {$(text.east)!.5!(latex.west)$}) |- (text.east);
\end{tikzpicture}%
\end{system}%


\begin{tikzpicture}[remember picture,overlay,line width=.3mm]%
  \coordinate (margin) at (.5\paperwidth-.5\textwidth, 0);
  \coordinate (topright) at (detikzify.north -| {$(current page.east)-(margin)$});
  \coordinate (topleft) at (detikzify.north -| {$(current page.west)+(margin)$});

  \node[anchor=north west] at (topleft) {%
    \scifig[%
      remember as=input1,%
    ]%
    {\usebox\pencil}
    {\scalebox{0.85}[1]{\includegraphics[height=1.1cm,trim=0 -40 0 -40]{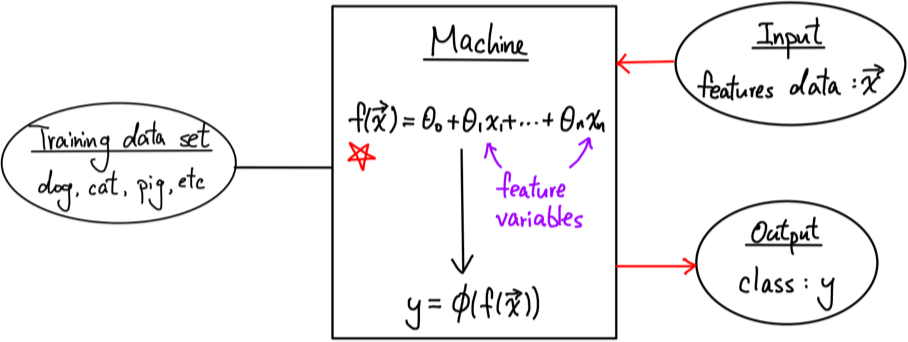}}}};
  \node[anchor=west] at (detikzify -| topleft) {%
    \scifig[%
      remember as=input2,%
    ]%
    {\usebox\camera}%
    {\includegraphics[height=1.1cm]{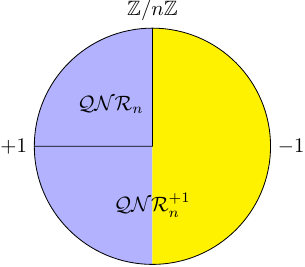}}};
  \node[anchor=south west] at (detikzify.south -| topleft) {%
    \scifig[%
      remember as=input3%
    ]%
    {\usebox\pencil}%
    {\includegraphics[height=1.1cm]{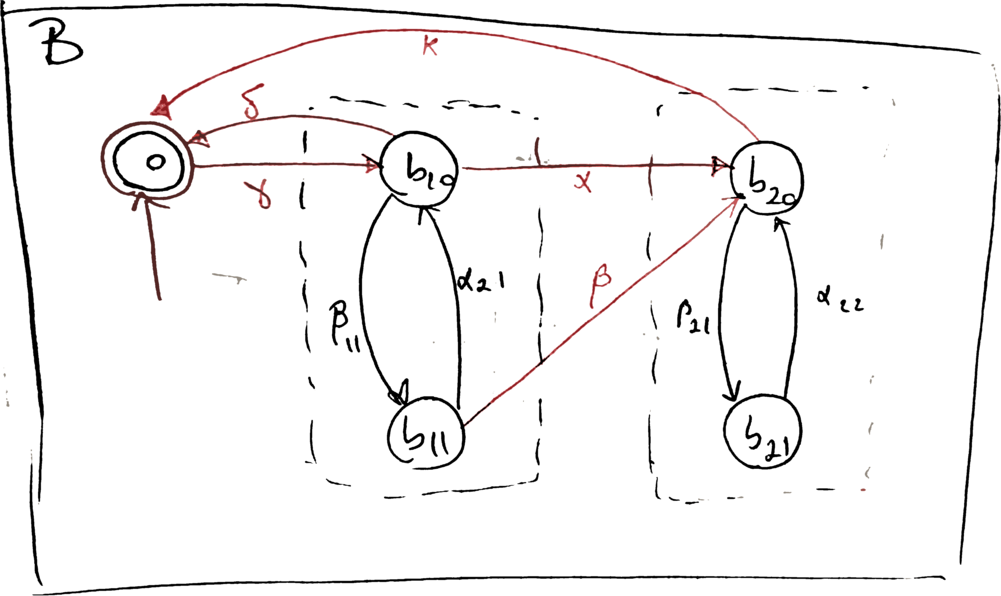}}};

  \foreach \i in {1,2,3} {
    \draw[line] (input\i.east) to [out=0,in=180] (input2.west -| {$(input1.east)!.5!(detikzify.west)$}) to
      [out=0,in=180] (detikzify.west |- vision) to  (vision.west);
  }

  \node[anchor=north east] at (topright) (output1) {%
    \scifig{}{\includegraphics[trim=0 -15 0 -15,height=1.1cm]{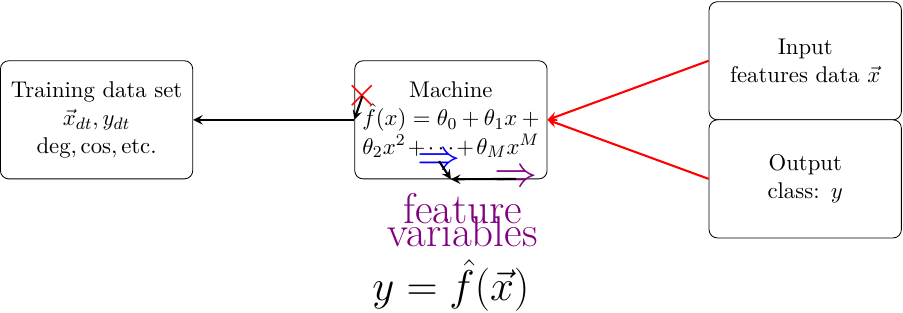}}%
  };
  \node[anchor=east] at (detikzify -| topright) (output2) {%
    \scifig{}{\includegraphics[height=1.1cm]{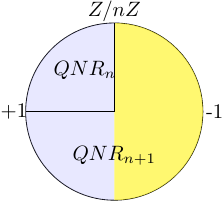}}%
  };
  \node[anchor=south east] at (detikzify.south -| topright) (output3) {%
    \scifig{}{\includegraphics[height=1.1cm]{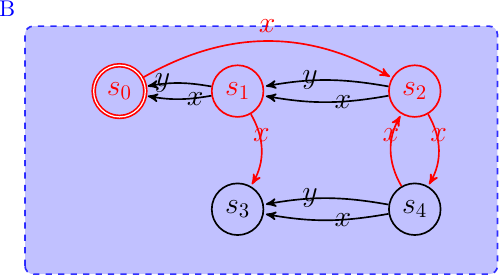}}%
  };

  \foreach \i in {1,2,3} {
    \draw[line] (latex.east |- detikzify.east) to (detikzify.east) to
      [out=0,in=180] (output\i.west);
  }
\end{tikzpicture}%
\endgroup%

%% file: sections/related.tex
\section{Related Work}\label{sec:related-work} 

\paragraph{Image-to-\latex Conversion}
A closely related task is the translation of mathematical illustrations into
\latex markup. In inspirational work, \citet{kirsch2010detexify} tackle the
recognition of single hand-drawn symbols to find corresponding \latex commands.
Subsequent works by
\citet{deng2017markup,zhang2017gru,zhang2019improved,wu2020math,wang2021latex}
expand on this concept to handle hand-drawn and scanned math formulas.
\citet{suzuki2003infty,wang2020pdf2latex,blecher2024nougat,lv2023kosmos25}
further extend the scope by extracting \latex formulas alongside text from
entire documents.

\paragraph{Image Vectorization}
Similarly, converting (rasterized) figures into \tikzname programs can be
characterized as a form of image
vectorization~\citep{sun2007vectorization,diebel2008bayes,ganin2018program,li2020diffvg,ma2022live,zhu2024samvg}.
Most existing methods vectorize images into low-level graphics primitives in
the \svg format~\citep{tian2024vector}. Although this works well for specific
domains like fonts, icons, and
emoji~\citep{lopes2019svgvae,carlier2020deepsvg,reddy2021im2vec,rodriguez2023starvector},
it does not capture higher-level semantics and does not generalize well to our
scientific context~(cf.\ \appref{sec:additional-results}).
Closer to our work, \citet{ellis2018drawing} generate vector representations as
graphics programs based on a limited subset of \latex commands. Their approach
even handles sketches, but their experiments are restricted to a synthetic
dataset with only basic shapes of limited complexity.
\citet{belouadi2024automatikz} also generate \tikzname programs, but their
primary emphasis is on conditioning the generation on textual descriptions,
with images serving only as a secondary input.

\paragraph{Code Generation}
As \tikzname is implemented in the Turing-complete \tex macro
system~\citep{erdweg2010tex}, our work is also closely tied to code
generation~\citep{xu2022codeeval}. Despite continuing progress in this
field~\citep{chen2021evaluating,li2022alphacode,li2023starcoder,guo2024deepseekcoder,lozhkov2024starcoder},
most research concentrates on high-resource languages like Python, Java, and
JavaScript~\citep{zan2023lnlcode}, typically overlooking \tex in evaluations.
However, \tex and \tikzname may still find their way into the training data, as
demonstrated by the zero-shot ability of some models to understand and generate
code in these
languages~\citep{bubeck2023sparks,belouadi2024automatikz,sharma2024vision}.

%% file: sections/data.tex
\section{Datasets}\label{sec:datasets} 
\begin{wraptable}[11]{R}{0pt}
  \centering
  \begin{tabular}{l s{3.3} s{3.3}}
    \toprule
    \thead{Source} & \nhead{\datikz[v1]} & \nhead{\datikz[v2]}\\
    \midrule
    curated    &    ,981 & 1,566\\
    \texse     &  29,238 & 30,609\\
    \arxiv     &  85,656 & 326,450\\
    artificial &   1,957 & 1,958\\
    \midrule
    all        & 117,832 & 360,583\\
    \bottomrule
  \end{tabular}
  \caption{Breakdown of the number of unique \tikzname graphics in \datikz[v2]
  compared to its predecessor \datikz[v1].}%
  \label{tab:dataset}
\end{wraptable}
We introduce \datikz[v2], to our knowledge, the most comprehensive dataset
of \tikzname graphics to date; \sketchfig, the first dataset comprising
human-created sketches of scientific figures; and \metafig, a large-scale
scientific figure dataset with rich metadata. See \appref{sec:examples} for
examples.

\paragraph{\datikz[v2]}
\datikz[v2] serves as the primary source of \tikzname graphics for training
\projectname. It is an expanded version of
\datikz[v1]~\citep{belouadi2024automatikz}, incorporating graphics from the
same sources, namely curated repositories, \texse, \arxiv papers, and
artificial examples. The key difference is that \datikz[v2] includes all
\tikzname programs that compile with
\texlive,\footnote{\url{https://tug.org/texlive}} regardless of whether they
have associated captions, which was a requirement for inclusion in \datikz[v1]
but is not needed for \projectname. This approach allows us to create a dataset
that is more than three times as large as its predecessor~(cf.\
\tabref{tab:dataset}).

\paragraph{\sketchfig}
To create realistic synthetic sketches of scientific figures in \datikz[v2], we
rely on examples of real human-created sketches.\ \texse is a suitable source
for collecting these, as users often illustrate their questions with sketches,
and the answers provide the desired figure. We semi-automatically extract these
figure-sketch pairs by first ranking all questions on the site that contain
images based on their similarity to the string \textquote{a sketch of a
scientific figure} using a multimodal vision encoder~\citep{zhai2023siglip}. We
retain the ones with high similarity scores, manually filter for
true positives, and align them with the best matching figure provided in the
answers. In total, we collect 549 figure-sketch pairs this way. As we also want
to use this dataset for evaluation~(cf.\ \secref{sec:experiments}), we ensure
that for a subset of these sketches, no code provided in the answers is
included in \datikz[v2].

\paragraph{\metafig}
Beyond \tikzname graphics, there is a much larger pool of figures where the
underlying source is not available. Existing datasets that collect such figures
frequently come with rich metadata, such as captions, OCR tokens, and
paragraphs that mention the
figures~\citep{hsu2021scicap,karishma2023aclfig,rodriguez2023ocrvqgan}. Since
such high-level descriptions are useful for pretraining~(cf.\
\secref{sec:models}; \citealp{liu2023llava}), we collect these datasets and
merge them with the subset of figures in \datikz[v2] that have captions. This
results in over 734k figure-text pairs, more than twice the size of
\datikz[v2].

%% file: sections/models.tex
\section{The \projectname Model}\label{sec:models} 
Building on previous
work~\citep{liu2023llava,liu2023improved,dai2023instructblip,mckinzie2024mm1},
we build \projectname by combining a pretrained vision encoder with a
pretrained language model~(cf.\ \figref{fig:detikzify}), where the vision
encoder receives figures or sketches as input images, and the language model
generates corresponding \tikzname programs as output. We focus on code language
models that have been pretrained on \tex, as this prior knowledge may be
helpful for our task. All the models we end up using follow the \llama
architecture~\citep{touvron2023llama}: \codellama~\citep{roziere2023code} has
likely been trained on \tex code from \arxiv~\citep{touvron2023llama}, as has
been \tinyllama~\citep{zhang2024tinyllama}, while \deepseek~(code variant;
\citealp{guo2024deepseekcoder}) was trained on \tex code from GitHub.
For the vision encoder, we use \siglip~\citep{zhai2023siglip}, which has been
trained on OCR annotations~\citep{chen2023pali} and demonstrates
state-of-the-art understanding of text-rich
images~\citep{tong2024eyes,chen2023pali3}, a crucial skill for our task. We
then condition the \llm{}s on \siglip's patch embedding vectors. To reduce the
prompt length, we concatenate adjacent patch
embeddings~\citep{chen2023minigptv2}. A feed-forward layer with dimensions
$2\delta_{\siglip}\times\delta_{\llm}$ serves as a connector, mapping image
features of dimension $\delta_{\siglip}$ to the \llm word embedding space of
dimension $\delta_{\llm}$.

\paragraph{Model Training}
We experiment with \tinyllama[1.1b] and \deepseek[1.3b] (approximately 1
billion parameters each) and \codellama[7b] and \deepseek[7b] (7 billion
parameters each). When referring to specific variants of \projectname, we use
the names \projectname[TL][1.1b], \projectname[DS][1.3b], \projectname[CL][7b],
and \projectname[DS][7b], respectively. For all models, we use the \sovit
variant of \siglip as the vision encoder.
Following \citet{liu2023llava,liu2023improved}, we first pretrain the connector
with other model parameters frozen. We pretrain for one epoch on \metafig with
\adam~\citep{loshchilov2018decoupled}, a batch size of 256, a learning rate of
\lr{1}{3}, and a cosine learning rate decay with a 3\% warmup ratio.
Next, we unfreeze the language model~(keeping the vision encoder frozen) and
fine-tune on examples from \datikz[v2] that fit within a 2048 token context
window. We use a batch size of 128, a learning rate of \lr{4}{5}, and train for
three epochs. Training data ablations can be found in
\appref{sec:additional-results}.

\paragraph{Synthetic Sketches}
When training \projectname on \datikz[v2], we randomly replace figures with
synthetic sketches 50\% of the time. Sketches are generated on the fly, meaning
that each time a figure is sampled as a sketch, a different synthetic sketch
will be generated.
Creating realistic sketches requires high-level image manipulation methods
that go beyond traditional transformations like zooming or cropping. We,
therefore, adopt \instructpix~\citep{brooks2023pix2pix}, a model capable of
diversely editing images based on human instructions. We chose this model due
to its remarkable zero-shot performance in generating synthetic sketches during
our initial experiments. By then fine-tuning the model on \sketchfig, we
further improve its performance~(cf.\ \secref{sec:analysis} and
\appref{sec:train-inf-details}).

%% file: sections/mcts.tex
\section{Iterative Refinement with \mcts*}\label{sec:mcts} 
Due to the inherent probabilistic nature of language models, generating valid
\tikzname programs during inference can be a challenging task. The generated
code may not always comply with the syntactic and semantic rules of \tex and
\tikzname, potentially leading to compilation errors. While constrained
decoding algorithms can assist in guiding models towards generating valid
programs~\citep{ugare2024improving,poesia2022synchromesh,scholak2021picard},
these approaches are limited to programming languages defined by context-free
grammars~(CFGs).\ However, \tex and \tikzname are not defined by
CFGs~\citep{erdweg2010tex}, rendering these methods ineffective for our
purpose. Moreover, even if the generated code compiles successfully, fidelity
errors such as misaligned elements, inconsistent scaling, repetitions, or
mislabeling may only become apparent in the rendered output.

Despite these challenges, which make it difficult to guide \projectname based
on intermediate states, we can still analyze completed outputs in a
straightforward manner~(e.g., by examining compiler diagnostics or comparing
rendered outputs to the input image), allowing us to make informed decisions
during subsequent sampling iterations. This concept of making decisions based
on random sampling of the search space forms the core of \mcts* (\mcts;\
\citealp{coulom2007mcts}). By integrating \projectname with \mcts and adapting
the standard \mcts algorithm to our problem domain, we can iteratively steer
\projectname towards more promising regions of the output space~(cf.\
\figref{fig:detikzify}). In the following, we outline our fundamental
approach, with further extensions discussed in \appref{sec:mcts-details}.

\subsection{Integrating \mcts into \projectname}
\mcts is a versatile search algorithm that has been successfully
applied to various domains, including board
games~\citep{silver2016go,silver2017go}, procedural content
generation~\citep{Kartal2016sokoban,Kartal2016bsokoban,Summerville2015mcmcts},
and more recently, guiding language models to achieve long-term
goals~\citep{brandfonbrener2024verified,zhang2023planning,chaffin2022ppl}.
The algorithm incrementally builds a search tree and repeatedly runs
simulations until an exit condition is met or a computational budget is
exhausted. In our context, at depth $n$, each node's state consists of $n$
lines of \tikzname code, and edges represent continuations for generating the
next line. Initially, \mcts starts with only an empty root node and then
iteratively performs the following four steps~(cf.\ \figref{fig:mcts}):
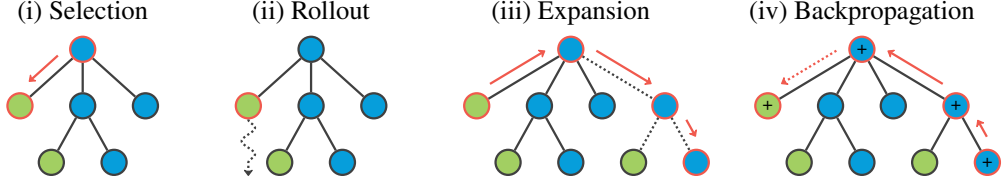
\begin{figure}
  \centering
  \input{mcts.tex}
  \caption{An example of the four steps of an \mcts simulation: The selection
  policy (i) reaches a green backtracking node (normal nodes are blue), causing
  new nodes from the rollout (ii) to be added to the parent node during
  expansion (iii). The reward is backpropagated (iv) accordingly.}%
  \label{fig:mcts}
\end{figure}

\paragraph{Selection}
Each simulation starts at the root node and successively selects child nodes
based on a \emph{selection policy} until a leaf node is reached. The policy
determines which parts of the tree should be explored further, balancing the
\emph{exploitation} of high-value regions and \emph{exploration} of less-visited
areas. Following previous work, we use Upper Confidence Trees (UCT;\
\citealp{kocsis2006bandit}) as our selection policy, iteratively selecting the
successor node $i$ that maximizes the formula
\begin{equation}
  \operatorname{UCT}(i) =
     \frac{\sum^{n_i}_{j=1}\dlmat[i,j]{V}}{\dlvec[i]{n}}
     +c\sqrt {\frac {\ln(\dlvec[\parent(i)]{n})}{\dlvec[i]{n}}},
\end{equation}
where $\dlmat[i,j]{V} \in [-1, 1]$ is the estimated value of $i$ at the $j$th
visit, $\dlvec[i]{n}$ and $\dlvec[\parent(i)]{n}$ are the visit counts
at $i$ and its parent $\parent(i)$, respectively, and $c$ is a coefficient that
controls the degree of exploration.

\paragraph{Rollout}
Once a leaf node is selected, we utilize \projectname as a \emph{rollout policy}.
By conditioning it on the node's state, we continue to sample \tikzname code
until the end-of-sequence token is encountered. This so-called rollout is then
stored for reuse in the subsequent steps.

\paragraph{Expansion}
Next, the tree is \emph{expanded} by adding nodes from the rollout as new leaf
nodes. While most implementations add only one node (i.e., one line of
\tikzname code) per simulation, computing rollouts with \llm{}s is
computationally expensive. Therefore, inspired by \mcts for real-time
settings~\citep{soemers2016realtime}, we instead add multiple nodes.
Specifically, we add $\sqrt{\norm{r}-d_l}$ new nodes, where
$\mathopen|r\mathclose|$ is the number of lines in rollout $r$ and $d_l$ is the
depth of the old leaf node $l$. This approach allows our tree to grow quickly
in early simulations while converging to the standard case in the long run.
To enable the tree to grow in multiple directions, we also introduce
\emph{backtracking}
nodes~\citep{brandfonbrener2024verified,chaslot2008progressive}. For each added
node $i$, we add a backtracking node as a sibling that mirrors the parent node
$\operatorname{p}(i)$. When a backtracking node is expanded, its descendants
are added to $\operatorname{p}(i)$ so that the backtracking node remains a
leaf. This enables a practically infinite search space anywhere in the tree
while still maintaining a bounded branching factor.

\paragraph{Backpropagation}
Finally, we calculate the value for rollout $r$ using a predefined reward
function~(cf.\ \secref{sec:mcts-rewards}) and \emph{backpropagate} it to every
node $i$ on the path from the root node to the newly added nodes by appending
it to $\dlmat[i,:]{V}$. We also increment the visit counts $n_i$ for the same
nodes. For backtracking nodes, only the visit counts are updated. Finally, we
check any exit conditions. If \mcts terminates, we return the \tikzname program
of the rollout that achieved the highest value.

\subsection{Reward Functions}\label{sec:mcts-rewards}
We explore two distinct reward functions to guide the search process. The first
reward function utilizes compiler diagnostics to identify documents that
compile successfully. The second reward function provides a visual signal based
on perceptual image similarity, which, in addition, helps find \tikzname
programs that better match the input image. We explore further reward functions
in \appref{sec:mcts-details}.

\paragraph{Compiler Diagnostics}\label{sec:diagnosstic-reward}
The diagnostics-based reward function is based on analyzing the log file from
compiling the generated \tikzname program. We assign rewards according to the
error state and whether an output file was produced. The reward function is
defined as follows:
\begin{equation}
  \dlmat[i,j]{V} =
    \begin{cases}
      \phantom{-}1 & \text{if the code compiles without issues},\\
      \phantom{-}0 & \text{if the code compiles with recoverable errors},\\
      -1 & \text{if compilation fails due to a fatal error}.
    \end{cases}
\end{equation}

\paragraph{Self-Assessed Perceptual Similarity (\ssim*)}
\ssim* computes the reward as the \emph{perceptual
similarity}~\citep{zhang2018lpips} between the input image and the compiled
output figure. We hypothesize that \projectname \emph{itself} can assess this
similarity, enabling the model to guide its own search process. To achieve
this, we encode both images into embedding vectors using \projectname's vision
encoder and calculate \ssim* as their cosine
similarity~\citep{fu2023dreamsim,hessel2021clipscore}. In cases where
compilation fails, we assign a reward of -1. In \secref{sec:analysis}, we
demonstrate that \ssim* correlates well with human judgments and outperforms
other baseline methods.

%% file: graphics/mcts.tex
\begingroup 

\pgfdeclaredecoration{simple line}{initial}{
  \state{initial}[width=\pgfdecoratedpathlength-1sp]{\pgfmoveto{\pgfpointorigin}}
  \state{final}{\pgflineto{\pgfpointorigin}}
}

\tikzset{
  backtrack/.style={fill=TUDa-4a},
  active/.style={draw=TUDa-9a},
  line/.style={draw=tcbcolframe,line width=.3mm},
  arrow/.style={line,->,>={Stealth[round,width=1.5mm,length=1mm]}},
  active arrow/.style={arrow,draw=TUDa-9a,shorten <=.75ex,shorten >=.75ex,decorate,decoration={simple line,raise=#1}},
  snake it/.style={decorate,decoration={snake,post length=1mm}},
}

\newcommand{\reward}{\makebox(0,0){\scriptsize+}}

\begin{forest}
  for tree={
    anchor=center,
    fill=TUDa-2a,
    circle,
    line,
    s sep*=2,
    l*=.5,
    edge path={%
      \noexpand\path[line,\forestoption{edge}] (!u.parent anchor)--(.child anchor);
      \noexpand\path[\forestoption{edge label}] (!u.parent anchor)--(.child anchor);
    }
  }
  [,phantom,s sep=1cm,
    [,active,name=selection
      [,backtrack,active,edge label={active arrow=-1ex}]
      [,
        [,backtrack]
        []
      ]
      []
    ]
    [,name=rollout
      [,backtrack,active,name=expand]
      [,
        [,backtrack]
        []
      ]
      []
    ]
    [,active,name=expansion
      [,backtrack,active,edge label={active arrow=-1ex,<-}]
      [,
        [,backtrack]
        []
      ]
      []
      [,active,edge={densely dotted},edge label={active arrow=1ex}
        [,backtrack,edge={densely dotted}]
        [,active,edge={densely dotted},edge label={active arrow=1ex},name=lowest]
      ]
    ]
    [\reward,active,name=backpropagation
      [\reward,active,backtrack, edge label={active arrow=-1ex, densely dotted}]
      [,
        [,backtrack]
        []
      ]
      []
      [\reward,active,edge label={active arrow=1ex,<-}
        [,backtrack]
        [\reward,active,edge label={active arrow=1ex,<-}]
      ]
    ]
  ]
  \draw[arrow,densely dotted,snake it] (expand.south) to (lowest.south -| expand.south);
  \node[anchor=south] at ($(selection)+(0,1ex)$) {(i) Selection\strut};
  \node[anchor=south] at ($(rollout)+(0,1ex)$) {(ii) Rollout\strut};
  \node[anchor=south] at ($(expansion)+(0,1ex)$) {(iii) Expansion\strut};
  \node[anchor=south] at ($(backpropagation)+(0,1ex)$) {(iv) Backpropagation\strut};
\end{forest}
\endgroup

%% file: sections/experiments.tex
\section{Experiments}\label{sec:experiments} 
Before training on \datikz[v2], we extract 1k samples to serve as our test
set for an automatic evaluation and generate corresponding synthetic sketches.
To mitigate data leakage from pretraining to testing, we only include items
created after the cut-off date of \codellama and exclude repositories that may
have been used in training \deepseek. We also use an \ngram matching algorithm
to prevent cross-contamination with our train split~\citep{openai2023gpt4}. For
a human evaluation involving human-created sketches, we also select 100
items from \sketchfig that do not overlap with \datikz[v2]~(cf.\
\secref{sec:datasets}). Across all models, we set the temperature to 0.8 and
the exploration coefficient $c$ to 0.6. We provide examples of real and
synthetic sketches as well as generated outputs in \appref{sec:examples} and
\tabref{tab:human-examples}.

\paragraph{Baselines}
Given \claude and \gpt's potential for our task (cf.\
\secref{sec:related-work}), we use them as baselines. Similar to \projectname,
we instruct these models to generate \tikzname programs for given images.
However, as proprietary chatbots, they often mix code and natural
language~\citep{zhang2023controllable,belouadi2024automatikz} and do not expose
the internals needed to compute \ssim*. This makes it impractical to apply our
\mcts-based refinement algorithm, which is designed for code-only outputs and
open models. Instead, we compare our approach to equivalent chat-oriented
refinement methods, i.e., we use Self-Refine as an alternative to
diagnostics-based \mcts and Visual Self-Refine as an alternative to
\ssim*-based \mcts~(\citealp{madaan2023selfrefine}; cf.\
\appref{sec:train-inf-details} for additional inference details). In
\appref{sec:additional-results}, we also explore \svg as an alternative to
\tikzname but find it less effective for our domain.

\subsection{Automatic Evaluation}\label{sec:auto-experiments}
We introduce two inference tasks to automatically evaluate our models on the
test split of \datikz[v2]. During \emph{output-driven} inference~(OI), we
employ the diagnostics-based reward and use successful compilation as an early
exit condition~(we consider compilation successful if an output artifact is
produced). For \emph{time-budgeted} inference~(TI), we use the more
fine-grained \ssim*-based reward and continue from OI until a computational
budget of 10 minutes is exhausted~(cf.\ \citealp{brandfonbrener2024verified}),
investigating the extent of achievable improvement. We report results for the
two use cases where either (rasterized) reference figures or (synthetic)
sketches serve as model inputs~(cf.\ \secref{sec:introduction}). Due to high
inference costs, we only evaluate commercial \claude and \gpt in OI using
Self-Refine, leaving TI with Visual Self-Refine for human evaluation. We
evaluate the following properties:
\begin{table*}
  \newcolumntype{\mtecol}[3]{>{\collectcell{\gradcol{#1}{91.221}{50.156}{#2}{#3}}}c<{\endcollectcell}}
  \newcolumntype{\cbleucol}[3]{>{\collectcell{\gradcol{#1}{1.815}{0.024}{#2}{#3}}}c<{\endcollectcell}}
  \newcolumntype{\tedcol}[3]{>{\collectcell{\gradcol{#1}{56.893}{60.298}{#2}{#3}}}c<{\endcollectcell}}
  \newcolumntype{\dsimcol}[3]{>{\collectcell{\gradcol{#1}{73.01}{59.102}{#2}{#3}}}c<{\endcollectcell}}
  \newcolumntype{\ssimcol}[3]{>{\collectcell{\gradcol{#1}{88.323}{73.954}{#2}{#3}}}c<{\endcollectcell}}
  \newcolumntype{\kidcol}[3]{>{\collectcell{\gradcol{#1}{5.951}{33.203}{#2}{#3}}}c<{\endcollectcell}}
  \newcolumntype{\avgcol}[3]{>{\collectcell{\gradcol{#1}{0.965}{0.148}{#2}{#3}}}c<{\endcollectcell}}

  \scriptsize
  \setlength{\extrarowheight}{\belowrulesep}\setlength{\belowrulesep}{0pt}
  \begin{widetabular}{\textwidth}{l
    \mtecol{2.3}{88.593}{88.03}  \cbleucol{1.3}{1.815}{1.477} \tedcol{2.3}{56.893}{57.178} \dsimcol{2.3}{73.01}{72.315}  \ssimcol{2.3}{88.323}{87.466} \kidcol{2.3}{5.951}{6.714}   \avgcol{1.3}{0.965}{0.869}
    \mtecol{2.3}{91.221}{90.597} \cbleucol{1.3}{0.69}{0.555}  \tedcol{2.3}{59.563}{59.693} \dsimcol{2.3}{65.198}{65.118} \ssimcol{2.3}{80.207}{79.717} \kidcol{2.3}{12.207}{17.334} \avgcol{1.3}{0.965}{0.941}}
    \toprule
    & \multicolumn{7}{c}{\thead{Reference Figures}} & \multicolumn{7}{c}{\thead{Synthetic Sketches}}\\
    \cmidrule(l{\tabcolsep}r{\tabcolsep}){2-8}\cmidrule(l{\tabcolsep}r{\tabcolsep}){9-15}
    \thead{Models} & \nhead{\mte\up} & \nhead{\cbleu\up} & \nhead{\ted\down} & \nhead{\dsim\up} & \nhead{\ssim\up} & \nhead{\kid\down} & \nhead{\underline{\avg}\up}
                   & \nhead{\mte\up} & \nhead{\cbleu\up} & \nhead{\ted\down} & \nhead{\dsim\up} & \nhead{\ssim\up} & \nhead{\kid\down} & \nhead{\underline{\avg}\up}\\
    \midrule
    \claude & 51.812 & 0.111 & 57.389 & 64.896 & 83.372 & 17.822 & 0.148
            & 50.156 & 0.024 & 59.731 & 59.102 & 73.954 & 29.541 & 0.189\\
    \gpt    & 61.975 & 0.286 & 57.178 & 69.741 & 86.215 &  6.714 & 0.612
            & 54.126 & 0.024 & 60.298 & 61.98  & 75.687 & 33.203 & 0.15\\[-\aboverulesep]
    \midrule
    \projectname*[TL][1.1b] & 88.03  & 1.168 & 58.815 & 65.538 & 84.161 & 15.747 & 0.207
                            & 90.597 & 0.502 & 60.202 & 60.585 & 77.947 & 21.851 & 0.454\\
    \projectname*[DS][1.3b] & 83.771 & 1.336 & 57.661 & 68.659 & 86.079 & 11.536 & 0.572
                            & 87.446 & 0.541 & 60.112 & 62.756 & 79.097 & 17.334 & 0.642\\
    \projectname*[CL][7b]   & 88.593 & 1.477 & 56.893 & 72.315 & 87.466 & 8.301  & 0.869
                            & 91.221 & 0.555 & 59.563 & 65.118 & 79.717 & 12.207 & 0.941\\
    \rule[\dimexpr-\dp\strutbox-\aboverulesep]{0pt}{0pt}%
    \underline{\projectname*[DS]}\unskip\textsubscript{7b}
                            & 82.366 & 1.815 & 57.227 & 73.01  & 88.323 &  5.951 & 0.965
                            & 89.299 & 0.69  & 59.693 & 65.198 & 80.207 & 12.207 & 0.965\\[-\aboverulesep]
    \bottomrule
  \end{widetabular}
  \caption{System-level scores for output-driven inference~(\projectname
    abbreviated as \projectname*). Bold and underlined values indicate the best
    and second-best scores for each metric column, respectively. Cell shading
    reflects the relative score magnitudes across input types. Arrows indicate metric
    directionality.}%
  \label{tab:fast-metrics}
\end{table*}
\begin{description}
  \item[Code Similarity] To measure the similarity between generated and
    reference \tikzname programs, we use \cbleu*~(\cbleu), a variant of \bleu
    optimized for evaluating code~\citep{eghbali2023bleu,papineni2002bleu}, and
    the \ted*~(\ted), our adapted version of the \eed*~\citep{stanchev2019eed}
    combined with a \tex tokenizer.
  \item[Image Similarity] In addition to \ssim*~(\ssim), which can also be used
    as a metric, we report \dsim*~(\dsim; \citealp{fu2023dreamsim}), a
    fine-tuned metric for perceptual similarity. We also compute the
    \kid*~($\text{\kid}\times10^3$; \citealp{bińkowski2018kid}), which assesses
    the overall quality of generated figures by comparing their distribution
    with the distribution of reference figures. These metrics are always
    computed by comparing the generated figures to the reference figures,
    regardless of what the model receives as input.
  \item[\avg*] To offer a holistic view of each model's performance, we also
    compute the arithmetic mean~(\avg) of all code and image similarity
    metrics. Given that these metrics operate on different scales, we min-max
    normalize their scores before calculating the average.
  \item[Efficiency] For OI, we compute the \mte*~(\mte) as the 10\% winsorized
    mean of the ratio of the number of tokens in the final \tikzname program to
    the total number of tokens generated to arrive at that program. For TI, we
    instead compute the \mst*~(\mst), measuring the throughput of unique
    \tikzname graphics for the given budget.
\end{description}

\paragraph{Results}
\tabref{tab:fast-metrics} presents the system-level metric scores for OI.\@ As
expected, the scores for reference figures are, on average, 38\% higher than
those for synthetic sketches, but similar patterns emerge across both input
types.\@ \projectname[CL][7b] and \projectname[DS][7b] consistently outperform
all other models, achieving \avg scores of 0.869 \& 0.965 for figures and 0.941
\& 0.965 for sketches, respectively. In contrast, \gpt reaches \avg scores of
only 0.612 and 0.15, placing it in competition with the smaller 1b models: for
figures, \gpt surpasses \projectname[TL][1.1b] and \projectname[DS][1.3b],
which achieve scores of 0.207 and 0.572, respectively. However, these smaller
models outperform \gpt on sketches, where they achieve scores of 0.454 and
0.642.\@ \claude trails behind all our models, with an \avg of only 0.148 and
0.189.
When examining individual similarity metrics, \projectname[ds][7b], the
top-performing \projectname model overall, surpasses \gpt, the best baseline,
by more than \pp{3} (\pp) on average for \dsim* and \ssim*, while maintaining a
noticeably lower \kid.
In terms of \cbleu, \gpt, and \claude only reach 6.5--18.5\% of the performance
achieved by the lowest-scoring \projectname model (\projectname[tl][1.1.b]).
The differences in \ted are less pronounced, possibly due to the influence of
boilerplate code, which \cbleu inherently ignores.

For efficiency, all \projectname models demonstrate an \mte of 82--91\%,
indicating that only 1--2 out of 10 inference runs require a second simulation
to generate a compilable \tikzname program. Interestingly, the model size does
not seem to particularly influence this score, with the pretraining setup
appearing to be the key factor instead. For instance, \projectname[tl][1.1b]
and \projectname[cl][7b] share a similar pretraining setup and exhibit
comparable \mte values, as do \projectname[ds][1.3b] and \projectname[ds][7b].
We can further observe that~(i) \mte is generally higher for sketches compared
to figures, and~(ii) for figures, the \mte of similarly pretrained models is
inversely correlated with their scores on other metrics. These phenomena likely
stem from models making fewer mistakes when the input is less detailed or when
their understanding of it is limited---a finding that aligns well with other
studies~\citep{tong2024eyes}. Compared to \projectname, \claude and \gpt
perform considerably worse, with an \mte of only 50--62\%. Notably, for these
models, 98.5\% of the items already compile after the initial Self-Refine step,
meaning that this inefficacy primarily originates from the natural language
texts surrounding the code and that Self-Refine is nearly equivalent to regular
sampling-based inference.

\begin{table*}
  \scriptsize
  \begin{widetabular}{\textwidth}{l *{5}{d{2.3}} d{3.3} d{1.3} *{6}{d{2.3}} d{1.3}}
    \toprule
    & \multicolumn{7}{c}{\thead{Reference Figures}} & \multicolumn{7}{c}{\thead{Synthetic Sketches}}\\
    \cmidrule(l{\tabcolsep}r{\tabcolsep}){2-8}\cmidrule(l{\tabcolsep}r{\tabcolsep}){9-15}
    \thead{Models} & \nhead{\mst\up} & \nhead{\cbleu\up} & \nhead{\ted\down}
    & \nhead{\dsim\up} & \nhead{\ssim\up} & \nhead{\kid\down}
    & \nhead{\underline{\avg}\up} & \nhead{\mst\up} & \nhead{\cbleu\up}
    & \nhead{\ted\down} & \nhead{\dsim\up} & \nhead{\ssim\up}
    & \nhead{\kid\down} & \nhead{\underline{\avg}\up}\\
    \midrule
    \projectname*[TL][1.1b] & \first{2.3}{33.775} & \wrs-0.011 & \btr-2.001 & \btr+8.704 & \btr+5.561 & \btr-12.146 & 0.128
                            & \first{2.3}{35.975} & \btr+0.094 & \btr-0.628 & \btr+5.82  & \btr+3.026 & \wrs+0.854 & 0.014\\
    \projectname*[DS][1.3b] & \second{2.3}{29.975} & \wrs-0.028 & \btr-1.303 & \btr+8.464 & \btr+5.108 & \first{3.3}{\btr-8.728} & 0.531
                            & \second{2.3}{32.429} & \btr+0.061 & \btr-0.504 & \btr+5.573 & \btr+2.685 & \wrs+5.493 & 0.22\\
    \projectname*[CL][7b]   & 25.124 & \second{2.3}{\btr+0.07}  &  \first{2.3}{\btr-1.351} & \first{2.3}{\btr+7.797} &  \first{2.3}{\btr+4.93}  & \second{3.3}{\btr-4.868} & \first{1.3}{0.876}
                            & 26.219 & \second{2.3}{\btr+0.073} & \second{2.3}{\btr-0.468} & \first{2.3}{\btr+5.079} & \second{2.3}{\btr+2.455} & \second{2.3}{\wrs+5.493} & \second{1.3}{0.681}\\
    \underline{\projectname*[DS]}\unskip\textsubscript{7b}
                            & 24.145 & \first{2.3}{\wrs-0.073} & \second{2.3}{\btr-1.542} & \second{2.3}{\btr+6.974} & \second{2.3}{\btr+3.893} & \btr-0.946 & \second{1.3}{0.76}
                            & 26.195 & \first{2.3}{\btr+0.054} & \first{2.3}{\btr-0.696}  & \second{2.3}{\btr+4.887} & \first{2.3}{\btr+2.241}  & \first{2.3}{\wrs+1.099} & \first{1.3}{0.994}\\
    \bottomrule
  \end{widetabular}
  \caption{System-level scores for time-budgeted inference, displaying relative
    changes for metrics shared with output-driven
    inference~(\tabref{tab:fast-metrics}; colored green for improvements and
    red for declines) and absolute scores for independent metrics. Bold and
    underlined values indicate the best and second-best \emph{absolute} scores
    for each metric column, respectively. Arrows indicate metric
    directionality.}%
  \label{tab:timeout-metrics}
\end{table*}
The results for \projectname on TI are presented in
\tabref{tab:timeout-metrics}. Remarkably, increasing the computational budget
for \mcts improves nearly all metrics for both reference figures and sketches
as input without requiring access to any additional knowledge. The improvement
with sketches is particularly noteworthy, as it demonstrates that the
refinement process enhances the desired properties even when the model input
type differs from the one used for evaluation. The \pp{2.2--5.6} increase of
\ssim* for all models is not surprising since it serves as the reward signal we
optimize, but \dsim* and \ted also increase by \pp{4.9--8.7} and \pp{0.5--2},
respectively, demonstrating the efficacy of our approach. While \kid improves
by 1--12.1 points with reference figures, it drops by 0.9--5.5 points with
sketches. We believe this is because sketches often omit minor details, such as
axis tick labels, which is reflected more in the output of the TI models,
biasing their overall output distributions. Therefore, we consider the
substantial improvement of metrics capturing instance-level similarities to be
more important. For \cbleu, we observe only minor changes~(less than
\pp{\textpm0.1}), aligning with findings that \bleu-based metrics become less
effective as performance increases~\citep{ma2019results}. The \mst and \avg
reveal that, although 1b models produce more unique outputs within the time
frame compared to their larger 7b counterparts~(30--36 vs. 24.1--26.2), they
still fail to close the overall gap in performance, with \avg scores ranging
between 0.014--0.531 compared to 0.681--0.994 for 7b models.

Overall, all \projectname models are capable of generating compilable outputs
with reasonable efficiency. Upon examination of these outputs, it becomes
evident that the 7b models, particularly \projectname[ds][7b], consistently
outperform both \claude and \gpt, whose performance is more comparable to the
1b range. Increasing the computational budget for \projectname further improves
performance.

\subsection{Human Evaluation}\label{sec:human-eval}
To further assess the quality of the generated figures, we perform a human
evaluation on \sketchfig using \emph{\bws*}~(\bws;
\citealp{louviere2015bws,kiritchenko2016bws,kiritchenko2017bws}). In this
process, for each reference figure, we present annotators with a tuple of
generated figures and ask them to identify the most and least perceptually
similar figure. We then transform this data into scores ranging from -1~(poor)
to 1~(excellent) by calculating the difference between the proportion of times
a figure is selected as the best and the proportion of times it is chosen as
the worst~\citep{orme2009maxdiffa}.
To keep the workload manageable, we focus on the most promising \projectname
model~(\projectname[ds][7b]) and the strongest baseline~(\gpt). Building upon
the automatic evaluation, we assess these models in the OI and TI
configurations, using either reference figures or human-created sketches as
input. For each input type, we engage six unique expert annotators~(cf.\
\appref{sec:demographics} for more details).

\begin{figure}[tb]
  \centering
  \includegraphics{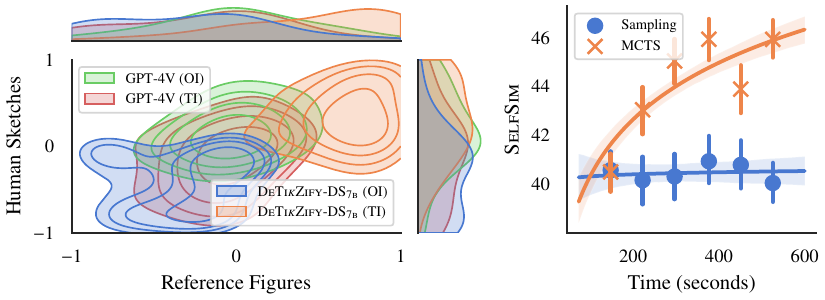}
  \caption{Bivariate distributions of \bws scores (higher is better) using
  kernel density estimation~(left) and log-linear regression over TI reward
  scores for different generation strategies over time~(right).}%
  \label{fig:bwc-strategy}
\end{figure}
\paragraph{Results}
\figref{fig:bwc-strategy}~(left) shows kernel density estimates for the
computed \bws scores, revealing intriguing findings that are consistent across
input types. In contrast to the automatic evaluation, \projectname[ds][7b]
performs worse (mean score $\mu=-0.32$) than \gpt~($\mu=0.09$) in OI.\@ This
could be attributed to the fact that \texse, the sole source of \sketchfig,
emphasizes minimum working examples, a type on which \gpt particularly
excels~\citep{belouadi2024automatikz}.
However, when we increase the computational budget, as in
\projectname[ds][7b]~(TI), it not only improves over OI results~($\mu=0.39$; in
line with automatic evaluation) but also surpasses \gpt in both configurations
by a considerable margin. Interestingly, \gpt's performance in TI ($\mu=-0.16$)
is lower than its performance in OI, indicating that \gpt~(TI) struggles to
refine its own outputs effectively and quickly deteriorates. Overall, this
shows how difficult it is for models to refine their own outputs and highlights
the effectiveness of our \mcts-based approach. Example outputs are provided in
\tabref{tab:human-examples}.
\begin{table*}
  \scriptsize
  \newlength\q
  \pgfmathsetlength{\q}{\textwidth/5 -2\tabcolsep}
  \newcommand{\includerow}[2][png]{%
    \includegraphics[height=1in,width=\q,keepaspectratio]{evaluation/human/#2/human_#1.#1} &
    \includegraphics[height=1in,width=\q,keepaspectratio]{evaluation/human/#2/gpt-4-textual_pdf.pdf} &
    \includegraphics[height=1in,width=\q,keepaspectratio]{evaluation/human/#2/gpt-4-visual_pdf.pdf} &
    \includegraphics[height=1in,width=\q,keepaspectratio]{evaluation/human/#2/detikzify-ds-7b-fast_pdf.pdf} &
    \includegraphics[height=1in,width=\q,keepaspectratio]{evaluation/human/#2/detikzify-ds-7b-timeout_pdf.pdf}\\
  }
  \begin{tabular}{*{5}{>{\centering\arraybackslash}m{\q}}}
    \toprule
    \thead{Input} & \thead{\gpt (OI)} & \thead{\gpt (TI)} & \thead{\projectname[DS][7b] (OI)} & \thead{\projectname[DS][7b] (TI)}\\
    \midrule
    \includerow{9}
    \midrule
    \includerow[pdf]{10}
    \midrule
    \includerow{11}
    \bottomrule
  \end{tabular}
  \caption{Examples of model inputs and generated outputs from our human
  evaluation, where annotators rated \gpt~(OI) higher than
  \projectname[DS][7b]~(OI) but ranked \projectname[DS][7b]~(TI) as the overall
  best model, illustrating our findings in \secref{sec:human-eval}. See
  \appref{sec:examples} for more examples.}%
  \label{tab:human-examples}
\end{table*}

%% file: sections/analysis.tex
\section{Analysis}\label{sec:analysis} 
In this section, we take a closer look at our methodologies and evaluation
strategies, correlating evaluation metrics with human judgments,
quantifying the quality of synthetic sketches, and examining the rate of
convergence of our \mcts algorithm. We also demonstrate that our models are not
affected by memorization of the training data, as shown in
\appref{sec:additional-results}.

\paragraph{Correlating Humans and Metrics}
\begin{wraptable}[10]{R}{0pt}
  \centering
  \begin{tabular}{l d{1.3} d{1.3}}
    \toprule
    \thead{Metric} & \nhead{Segment} & \nhead{System}\\
    \midrule
    \lpips &              0.224  & \second{1.3}{0.642}\\
    \dists &              0.32   & \second{1.3}{0.642}\\
    \dsim  & \second{1.3}{0.424} &  \first{1.3}{0.954}\\
    \ssim  &  \first{1.3}{0.436} & \second{1.3}{0.642}\\
    \bottomrule
  \end{tabular}
  \caption{Correlations of image similarity metrics with humans at the segment
  and system level.}%
  \label{tab:correlation}
\end{wraptable}
To assess the reliability of our human evaluation results, we investigate the
agreement between annotators. To this end, we calculate the \emph{\shr*}~(\shr;
\citealp{kiritchenko2017bws}) by randomly splitting our annotations into two
subsets, computing \bws scores for each subset, and measuring their correlation
with Spearman's $\rho$. The \shr values of 0.69 for sketches and 0.75 for
images indicate a moderate to strong correlation between annotators,
supporting the validity of our human evaluation results.
Motivated by these findings, we explore whether metrics that also assess
perceptual similarity~(i.e., \ssim* and \dsim*) correlate with these human
judgments. We again calculate Spearman's $\rho$ and show the average
correlations~\citep{corey1998averaging} at the segment and system level in
\tabref{tab:correlation}. For comparison, we also include the popular \lpips
and \dists metrics~\citep{zhang2018lpips,ding2022dists}. At the segment level,
\ssim* outperforms all other metrics, which is remarkable considering it is the
only untrained metric. Segment-level performance is particularly important for
fine-grained reward functions, justifying our choice of \ssim* in our \mcts
algorithm. At the system level, \dsim* performs the best, showcasing its
strength in evaluation settings.

\paragraph{Synthetic Sketch Quality}
We also assess the quality of our synthetic sketches by measuring their
congruence coefficient~\citep{sava2006congruence} with real sketches. We embed
human-created figure-sketch pairs from \sketchfig using \siglip, subtract each
sketch embedding from the corresponding figure embedding to obtain \emph{local}
sketch vectors, and perform a single-component Principal Component Analysis to
derive a \emph{global} sketch vector~\citep{zou2023representation}. We repeat
this process for synthetic sketches generated for the test split of \datikz[v2]
and compare the global vectors using cosine similarity. Base \instructpix
generates synthetic sketches with a congruence coefficient of 0.66, which
increases to 0.7 after fine-tuning. These results demonstrate a high
correlation with human-created sketches, suggesting that our generated sketches
are of good quality.

\paragraph{\mcts Convergence}
To gain insights into the long-term characteristics of our \mcts algorithm, we
visualize the trends in achieved TI reward scores over time in
\figref{fig:bwc-strategy}~(right) and compare them to conventional
sampling-based inference. As expected, sampling does not lead to improvements
over time due to the absence of a feedback loop. In contrast, \mcts
consistently improves throughout the entire time frame, and even at the end of
our budget of 10 minutes, it does not appear to converge, suggesting potential
additional gains for larger budgets. Apart from this, \mcts is not only more
effective but also faster. With an average \mst of 25.17, compared to 18.7 for
sampling, our \mcts algorithm generates considerably more unique \tikzname
programs within the same amount of time.

%% file: sections/conclusion.tex
\section{Conclusion} 
In this work, we showcase the potential of \projectname in generating \tikzname
programs for two practical use cases. First, it can convert existing figures
from lower-level formats into \tikzname, paving the way for semantic image
editing and downstream tasks~\citep{zhang2023custom}. Second, it can develop
hand-drawn sketches into \tikzname graphics, which could aid researchers in
creating high-quality scientific illustrations. In both cases, \projectname
substantially outperforms the commercial \llm{}s \gpt and \claude despite its
presumably much smaller size. We hope that our datasets~(\datikz[v2],
\sketchfig, and \metafig), our method for generating synthetic sketches, and
our \mcts-based inference algorithm will pave the way towards future research
on graphics program synthesis and bolster the cause of open science.

Looking ahead, we plan to extend our approach to other graphics languages, such
as MetaPost, PSTricks or
Asymptote~\Citep{hobby2014metapost,zandt2007pstricks,asymptote2024hammerlindl}.
We also intend to explore alternatives to perceptual similarity as an \mcts
reward signal, including per-pixel measures and point cloud
metrics~\citep{wang2009mse,wu2021balanced}. In addition, we aim to investigate
reinforcement learning from reward functions, for example, using Direct
Preference Optimization~\citep{rafailov2023direct,xu2024contrastive}. Finally,
while this work focuses on visual inputs, we plan to explore additional
modalities, such as text and mixed-modality inputs, in future work.

%% file: sections/limitations.tex
\section*{Limitations}\label{sec:limitations} 
In this work, we compare openly available models with proprietary systems that
lack transparency in their training details and internal workings and whose
performance is not stable over time. This inevitably complicates efforts to
address concerns such as data leakage or cross-contamination and limits the
fairness and reproducibility of our experiments. Nevertheless, under these
adverse conditions, our open models and methods demonstrate favorable
performance. Users should be aware, however, that our models might inherit
biases, flaws, or other limitations present in the training data, potentially
leading to discrepancies between expected results and generated outputs.
Furthermore, given the resource-intensive nature of \llm{}s, many of our
training and inference hyper-parameters were adopted from related work or
chosen based on general intuition. Although \llm{}s are generally robust to
hyper-parameter selection \citep{beyer2024paligemma}, conducting a thorough
hyper-parameter search might enhance their performance further. Finally, it
should be noted that our models could potentially be misused by malicious
actors to produce misinformation and fake science.

Another important consideration is that the public release of \datikz[v2] does
not include some \tikzname programs from our internal version due to licensing
restrictions. These programs are distributed under the \arxiv.org perpetual,
non-exclusive license, which prohibits redistribution. Nonetheless, we provide
our dataset creation scripts alongside usage instructions, enabling anyone to
reproduce the full version of \datikz[v2] independently. The remaining
\tikzname programs in \datikz[v2] are licensed under Creative Commons
attribution licenses,\footnote{\url{https://creativecommons.org/licenses}} the
GNU Free Documentation
License,\footnote{\url{https://www.gnu.org/licenses/fdl-1.3.en.html}} or the
MIT license,\footnote{\url{https://opensource.org/license/mit}} and their
respective terms and conditions apply. Regarding artificially created examples,
\citeauthor{openai2023gpt4}'s terms of use restrict the use of their services
for creating competing products, limiting this subset of \datikz[v2] to
non-commercial
applications.\footnote{\url{https://openai.com/policies/terms-of-use}}

%% file: sections/acknowledgements.tex
\section*{Acknowledgments}
We would like to express our sincere gratitude to the following individuals for
their contributions to our work: JiWoo Kim, Tommaso Green, Christoph Leiter,
Ines Reinig, Martin Kerscher, Margret Keuper, Christopher Klamm, Daniil
Larionov, Yanran Chen, Tornike Tsereteli, and Daniel Ruffinelli. Their
assistance with our human evaluation campaign, proofreading, insightful
discussions, and constructive comments have been invaluable. The last author is
supported by the Federal Ministry of Education and Research~(BMBF) via the
research grant \textsc{Metrics4NLG} and the German Research Foundation~(DFG)
via the Heisenberg Grant EG~375/5--1. We would also like to acknowledge the
OpenMoji project for providing the open-source icons used throughout this work
and Hugging Face for their generous community GPU grant.

%% file: sections/appendix.tex
\section{Further Details on \mcts}\label{sec:mcts-details}  
In this section, we discuss several extensions to our \mcts algorithm that aim
to improve its performance and efficiency. We also explain alternative reward
functions that we experimented with but ultimately found less effective
than our chosen approaches.

\subsection{\mcts Enhancements}\label{sec:mcts-enhancements}
Building on our base \mcts implementation, we introduce several enhancements,
namely dynamic rescaling of visual rewards, node deduplication, and preemptive
stopping of faulty rollouts.

\paragraph{Dynamic Rescaling}
One challenge when using \ssim* is that \mcts expects values to be in
the range of $[-1,1]$, while deep encoders often work with a much narrower
range in practice~\citep{hessel2021clipscore,zhang2020bertscore}. Furthermore,
this range may vary depending on whether the input image is a real figure or a
sketch. To address this discrepancy, we propose dynamically min-max normalizing
the visual reward scores whenever they are (re)computed, ensuring 
that \mcts always operates on the full range. The modified reward formula is as
follows:
\begin{equation}
  \dlmat[i,j]{V'} =
  \begin{cases}
    \frac
      {\dlmat[i,j]{V}-\min(\wout{\dlmat[i,:]{V}}{\{-1\}})}
      {\max(\dlmat[i,:]{V})-\min(\wout{\dlmat[i,:]{V}}{\{-1\}})}
     & \text{if $\dlmat[i,j]{V}\neq-1$
       and $\max(\dlmat[i,:]{V})\neq\min(\wout{\dlmat[i,:]{V}}{\{-1\}})$},\\
    \phantom{-}0 & \text{if $\dlmat[i,j]{V}\neq-1$
       and $\max(\dlmat[i,:]{V})=\min(\wout{\dlmat[i,:]{V}}{\{-1\}})$},\\
    -1 & \text{otherwise}.
  \end{cases}
\end{equation}

\paragraph{Node Deduplication}
During a rollout for a backtracking node, it is possible to generate code that
already exists elsewhere in the tree~(i.e., in siblings and their descendants).
To prevent the duplication of nodes, we always merge identical
node states before adding any nodes to the tree.

\paragraph{Preemptive Stopping}
If the code generated in a rollout cannot be compiled due to a fatal error, we
record the rollout, including the state in which the faulty line of code was
first introduced. If the same (intermediate) state is sampled again during
subsequent rollouts, we know that the completed output will fail to compile. In
such cases, we preemptively abort the rollout and reuse the previously recorded
rollout for the remainder of the simulation. To further prevent continuations
from faulty code, during the expansion phase, we only add nodes to our tree
whose node states do not contain any lines of code with fatal errors.

\subsection{Additional Reward Functions}\label{sec:additional-rewards}
Taking inspiration from popular machine translation
metrics~\citep{belouadi2023uscore,zhao2019moverscore,zhao2020limitations,song2021sentsim},
which compute the \emd*~(\emd; \citealp{rubner1998emd,kusner2015wmd}) between
word embeddings, we also explore with measuring perceptual image similarity
as the \emd between \siglip's image patch embeddings. Given the distance matrix
$\dlmat{D}$, where $\dlmat[i,j]{D}=\cos(\dlvec[i]{x}, \dlvec[j]{y})$ and
$\dlvec{x},\dlvec{y}$ are the patch embedding vectors of the input and output
images of simulation $j$ with lengths $\norm{\dlvec{x}}$ and
$\norm{\dlvec{y}}$, respectively, \emd is defined as follows:
\begin{equation}
  \operatorname{EMD}(x,y)=\frac
    {\sum _{i=1}^{\norm{\dlvec{x}}}\sum _{j=1}^{\norm{\dlvec{y}}}\dlmat[i,j]{F}\dlmat[i,j]{D}}
    {\sum _{i=1}^{\norm{\dlvec{x}}}\sum _{j=1}^{\norm{\dlvec{y}}}\dlmat[i,j]{F}},
  \quad\text{with}\quad
    \min\limits_{\dlmat{F}\geq0}{\sum_{i=1}^{\norm{\dlvec{x}}}\sum_{j=1}^{\norm{\dlvec{y}}}\dlmat[i,j]{F}\dlmat[i,j]{D}}
  \quad\text{s.t.}\quad\forall_{i,j}
    \left\lbrace\def\arraystretch{1.2}\begin{array}{@{}l@{}l@{}}
      \sum_{i=1}^{\norm{\dlvec{x}}}\dlmat[i,j]{F} &= \frac{1}{\norm{\dlvec{y}}},\\
       \sum_{j=1}^{\norm{\dlvec{y}}}\dlmat[i,j]{F} &= \frac{1}{\norm{\dlvec{x}}}.
     \end{array}\right.
\end{equation}
We define $\dlmat[i,j]{V}=2\tanh(-\operatorname{EMD(x,y)})+1\in[-1,1]$ if
compilation produces any output. If compilation fails, we set the reward to -1.
We empirically tune the hyperparameter on which layer to extract the patch
embeddings using the perceptual similarity dataset of scientific figures from
\citet{belouadi2024automatikz}. We find that extracting embeddings after the
24th layer yields the best results.
However, when evaluated on our data, this reward function achieves a
segment-level correlation of only 0.425~(cf.\ \secref{sec:analysis}), which is
lower than for \ssim* while being computationally more expensive. Consequently,
we do not employ this reward function in further experiments.

\section{Additional Experimental Results \& Analyses}\label{sec:additional-results}
\begin{wrapfigure}[20]{R}{0pt}
  \includegraphics{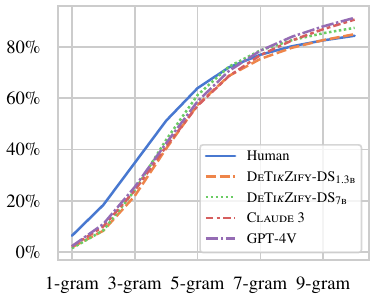}
  \caption{Proportion of generated code \ngram{}s with $n \in [1,10]$ that are
  novel~(i.e., not present in the training data). Results for human-created
  code are included as a reference point for comparison.}%
  \label{fig:memorization}
\end{wrapfigure}

In \tabref{tab:tikz-live}, we compare \live~\citep{ma2022live}, a
state-of-the-art method for generating \svg, with our \tikzname-based approach.
In \figref{fig:memorization}, we additionally investigate the extent to which
our models memorize the training data. We also perform training data ablation
studies, as presented in \tabref{tab:ablation}.
\begin{table*}
  \centering
  \newcolumntype{\dsimcol}[2]{>{\collectcell{\gradcol{2.3}{73.01}{49.455}{#1}{#2}}}c<{\endcollectcell}}
  \newcolumntype{\ssimcol}[2]{>{\collectcell{\gradcol{2.3}{88.323}{64.998}{#1}{#2}}}c<{\endcollectcell}}
  \newcolumntype{\kidcol}[2]{>{\collectcell{\gradcol{3.3}{5.951}{416.016}{#1}{#2}}}c<{\endcollectcell}}

  \setlength{\extrarowheight}{\belowrulesep}\setlength{\belowrulesep}{0pt}
  \begin{tabular}{l \dsimcol{73.01}{72.315} \ssimcol{88.323}{87.466} \kidcol{5.951}{6.714} \dsimcol{65.198}{65.118} \ssimcol{80.207}{79.717} \kidcol{12.207}{17.334}}
    \toprule
    & \multicolumn{3}{c}{\thead{Reference Figures}} & \multicolumn{3}{c}{\thead{Synthetic Sketches}}\\
    \cmidrule(lr){2-4}\cmidrule(lr){5-7}
    \thead{Models} & \nhead{\dsim\up} & \nhead{\ssim\up} & \nhead{\kid\down} &
    \nhead{\dsim\up} & \nhead{\ssim\up} & \nhead{\kid\down}\\
    \midrule
    \live   & 57.078 & 69.253 & 324.219 & 49.455 & 64.998 & 416.016\\[-\aboverulesep]
    \midrule
    \claude & 64.896 & 83.372 &  17.822 & 59.102 & 73.954 & 29.541\\
    \gpt    & 69.741 & 86.215 &   6.714 & 61.98  & 75.687 & 33.203\\[-\aboverulesep]
    \midrule
    \projectname[TL][1.1b] & 65.538 & 84.161 & 15.747 & 60.585 & 77.947 & 21.851\\
    \projectname[DS][1.3b] & 68.659 & 86.079 & 11.536 & 62.756 & 79.097 & 17.334\\
    \projectname[CL][7b]   & 72.315 & 87.466 &  8.301 & 65.118 & 79.717 & 12.207\\
    \rule[\dimexpr-\dp\strutbox-\aboverulesep]{0pt}{0pt}%
    \projectname[DS][7b]   & 73.01  & 88.323 &  5.951 & 65.198 & 80.207 & 12.207\\[-\aboverulesep]
    \bottomrule
  \end{tabular}
  \caption{System-level scores for \live, an \svg-generating model, compared
    with \tikzname-based models from output-driven inference. Scores for
    \tikzname-based models are copied from \tabref{tab:fast-metrics} for easy
    reference. Bold and underlined values indicate the best and second-best
    scores for each metric column, respectively. Cell shading reflects the
    relative score magnitudes across input types. Arrows indicate metric
    directionality.}%
  \label{tab:tikz-live}
\end{table*}

\subsection{Comparing \tikzname and \svg}\label{sec:svg}
Since \live generates \svg code instead of \tikzname, we do not report \cbleu
and \ted scores. Additionally, because it optimizes B\'ezier curves rather than
generating tokens, we exclude \mte, leaving only the image similarity metrics
\dsim*, \ssim*, and \kid. \tabref{tab:fast-metrics} shows that \live
underperforms all other models in our evaluation. On reference figures, it
scores over \pp{7.8} and \pp{14.1} lower than the worst baseline model on
\dsim* and \ssim*, respectively, and its \kid is more than 18 times higher.
This subpar performance can be attributed to the complexity of scientific
figures saved as \svg{}s. While we use \live in its default configuration,
generating eight paths with four segments each, our scientific figures consist
of over 110 paths on average with an arbitrary number of segments, not counting
deduplicated paths, which \live cannot detect. Although we could theoretically
configure \live to generate more paths, this would linearly increase inference
time, quickly becoming intractable.\ \live already requires over 18 hours to
complete the test set for one input type, whereas \projectname[CS][7b]~(OI),
for example, takes less than 5 hours. Furthermore, since \live attempts to
vectorize the input directly without semantic interpretation, it performs even
worse on synthetic sketches. We conclude that \svg, and, by extension, models
that generate \svg{}, are not well-suited for our problem domain and
objectives.

\subsection{Memorization}\label{sec:memorization}
Memorization of training data is a common concern in language
models~\citep{mccoy2023raven,carlini2023quantifying,raunak2022memo,meehan2020test}.
To assess the extent of this issue in our models, we calculate the \emph{\ngram
novelty}~\citep{mccoy2023raven}. Specifically, we determine the proportion of
\ngram{}s, with $n \in [1,10]$, in the model-generated \tikzname programs that
are \emph{not} present in the training data. We perform this analysis on the
test split of \datikz[v2] for our baselines and \deepseek-based \projectname
models conditioned on reference figures, as well as human-generated code, as
shown in \figref{fig:memorization}.
All models initially exhibit similar novelty and are slightly less novel than
humans for $n<7$. However, starting from $n=7$, all models except
\projectname[DS][1.3b] surpass human novelty, with more than 80\% of all
model-generated \ngram{}s being novel for $n>=8$. This phenomenon of models
becoming more novel than humans is commonly observed and is considered an
indicator that language models are not significantly affected by
memorization~\citep{mccoy2023raven,belouadi2024automatikz}.
Interestingly, for larger \ngram{}s, \projectname[DS][7b] demonstrates higher
novelty than its smaller counterpart, suggesting that despite its larger
capacity, it does not overfit and generalizes well. The most novel models are
\gpt and \claude, possibly because they were not trained on \datikz[v2] and
might have been trained on data that has been prepared differently.

\subsection{Training Data Ablation Studies}
To better understand the impact of training with synthetic sketches and
pretraining using \metafig on test set performance, we conducted ablation
studies with \projectname[DS][1.3b] in the OI configuration as a representative
model, following the experimental setup detailed in
\secref{sec:auto-experiments}.
\begin{table*}
  \centering\scriptsize
  \newcommand{\minus}[1]{$-\textrm{#1}$}
  \begin{widetabular}{\textwidth}{l *{6}{d{2.3}} d{3.3} *{4}{d{2.3}} d{3.3}}
    \toprule
    & \multicolumn{6}{c}{\thead{Reference Figures}} & \multicolumn{6}{c}{\thead{Synthetic Sketches}}\\
    \cmidrule(l{\tabcolsep}r{\tabcolsep}){2-7}\cmidrule(l{\tabcolsep}r{\tabcolsep}){8-13}
    \thead{Models} & \nhead{\mte\up} & \nhead{\cbleu\up} & \nhead{\ted\down}
    & \nhead{\dsim\up} & \nhead{\ssim\up} & \nhead{\kid\down}
    & \nhead{\mte\up} & \nhead{\cbleu\up} & \nhead{\ted\down}
    & \nhead{\dsim\up} & \nhead{\ssim\up} & \nhead{\kid\down}\\
    \midrule
    Full Training & 83.771 & 1.336 & 57.661 & 68.659 & 86.079 & 11.536
                  & 87.446 & 0.541 & 60.112 & 62.756 & 79.097 & 17.334\\
    \midrule
    \minus{Synthetic Sketches} & \wrs-1.957 & \btr+0.327 & \btr-0.822 & \btr+2.433 &
      \btr+1.318 & \wrs+3.296 & \wrs-13.358 & \btr+0.171 & \wrs+1.369 &
      \wrs-3.993 & \wrs-3.332 & \wrs+34.18\\
    \minus{\metafig} & \wrs-1.846 & \wrs-0.096 & \btr-0.356 & \btr+0.398 &
      \wrs-0.046 & \wrs+0.115 & \wrs-0.132 & \btr+0.053 & \btr-0.378 & \btr+0.084 &
      \wrs-0.181 & \wrs+2.773\\
    \bottomrule
  \end{widetabular}
  \caption{ Ablation study results for \projectname*[TL][1.1b] (OI), showing
  the relative impact on test set performance when either sketch-based training
  or connector pretraining is omitted, compared to full training. Improvements
  are highlighted in green, and declines in red, with reference scores taken
  from \tabref{tab:fast-metrics}.}\label{tab:ablation}
\end{table*}
In particular, \tabref{tab:ablation} compares full training with variations
where synthetic sketches are excluded and the step of pretraining the connector
is omitted. The results from excluding synthetic sketches align with
expectations: although this approach slightly improves performance on reference
figures on average, it substantially reduces performance on sketches.
Therefore, for models expected to perform well on both figures and sketches, we
recommend our original training methodology. Conversely, for models focused
solely on figures, training exclusively on figures may be advantageous.
The findings related to skipping connector pretraining are less definitive
as the score differences are minimal, reflecting the lack of consensus in
related literature about the benefits of connector pretraining for downstream
performance~\citep{liu2023llava,liu2023improved,karamcheti2024prismatic}.
However, on average, we observe a positive impact, especially on \mte and \kid,
where consistent improvements are noted for both reference figures and
synthetic sketches as input. Thus, we advocate incorporating a dedicated
pretraining step in the training protocol. In future work, we also plan
to investigate the impact of pretraining dataset size and quality.

\section{Additional Training \& Inference Details}\label{sec:train-inf-details}
In this section, we provide supplementary information on the training and
inference procedures for all our models. For training and inference of our
local \projectname models, we utilize a compute node equipped with four Nvidia
A40 GPUs and 448 gigabytes of RAM.\@ We access \claude and \gpt through their
respective official API endpoints.

\subsection{\projectname}
Complementing the information provided in \secref{sec:models}, our 1b models
require approximately two days of fine-tuning on our hardware. For the 7b
models, we employ optimizer state and gradient
partitioning~\citep{Rajbhandari2020zero} to accommodate them within the
available resources, resulting in an extended training time of 21 days.
Generating sketches for the training runs takes an additional 1.5 days, but
since we cache our sketches, these costs are incurred only once. Output-driven
inference takes 4--8 hours, depending on the model and input type, and
time-budgeted inference extends the runtime by a further 1.5 days.

\subsection{\instructpix}
As \sketchfig with only 549 examples may be considered too small for
fine-tuning \instructpix, we augment our training data with 4000 additional
sketches of natural images~\citep{sangkloy2016sketch,li2019sketch} and 2000
synthetic sketches of scientific figures generated with base \instructpix. We
then oversample \sketchfig at a 5:1 ratio and, following
\citet{paul2023instruct}, train for 15k steps with a batch size of 8 and a
learning rate of \lr{5}{5}. We select \textquote{turn it into a doodle} as our
initial prompt, which also appears in \instructpix's pretraining dataset and
demonstrates the most promising zero-shot performance.

\subsection{\claude \& \gpt}
Building upon the experiments described in \secref{sec:experiments}, we derive
all our Self-Refine prompts from the official examples provided by
\citet{madaan2023selfrefine} for generating \tikzname programs, with only minor
modifications. In particular, we employ the following prompt template in the
initial step of both Self-Refine and Visual Self-Refine, substituting
\textquote{sketch} or \textquote{picture} as appropriate:
\begin{prompt}
  This is a [ sketch | picture ] of a scientific figure. Generate LaTeX code
  that draws this scientific figure using TikZ. Ensure that the LaTeX code is
  self-contained and does not require any packages except TikZ-related imports.
  Don\textquotesingle{}t forget to include \textbackslash{}usepackage\{tikz\}!
  I understand that this is a challenging task, so do your best. Return your
  result in a \textasciigrave\textasciigrave\textasciigrave{}latex code block.
\end{prompt}
We then extract the first \latex code block from the generated text. In the
rare cases where \gpt incorrectly classifies input images as unsafe, we add a
small amount of Gaussian noise to the image pixels to bypass the issue. If
compilation fails due to a fatal error~(which occurs in only 1.5\% of all
cases) without producing an output artifact, we repeatedly use the following
prompt template until all issues are resolved, replacing \texttt{<code>} with
the generated code and \texttt{<error>} with the corresponding error message:
\begin{prompt}
  Given the error message:\\
  <error>\\
  And the problematic code:\\
  \textasciigrave\textasciigrave\textasciigrave{}latex\\
  <code>\\
  \textasciigrave\textasciigrave\textasciigrave{}\\
  First, identify the issue based on the error message. Then, determine the
  cause of the error in the code. Finally, propose and implement a solution.
  Return the fixed code in a
  \textasciigrave\textasciigrave\textasciigrave{}latex code block.
\end{prompt}
For Visual Self-Refine, we additionally use the following prompt template to
visually refine the output. Since we provide two input images~(the initial
figure or sketch and the current output), we label one as \textquote{Input} and
the other as \textquote{Reference}.\@ \claude's API has a built-in mechanism
for labeling images, while for \gpt, we embed the labels directly into the
images:
\begin{prompt}
  \textasciigrave\textasciigrave\textasciigrave{}latex\\
  <code>\\
  \textasciigrave\textasciigrave\textasciigrave{}\\
  This is the TikZ/LaTeX code for the scientific figure shown in the picture
  labeled \textquotedbl{}Input\textquotedbl. Can you improve it to better
  resemble the provided reference [ sketch | picture ]? First, analyze the
  \textquotedbl{}Input\textquotedbl{} picture to understand its components and
  layout. Then, consider how the scientific figure can be enhanced to more
  closely match the reference [ sketch | picture ]. Finally, rewrite the TikZ
  code to implement these improvements, making the image more similar to the
  reference. Ensure that the LaTeX code is self-contained and does not require
  any packages except TikZ-related imports. Don\textquotesingle{}t forget to
  include \textbackslash{}usepackage\{tikz\}! Return your result in a
  \textasciigrave\textasciigrave\textasciigrave{}latex code block.
\end{prompt}
Following the findings of \citet{madaan2023selfrefine}, we visually refine for
a maximum of four iterations, as they observe diminishing returns beyond that
point, and it helps reduce inference costs. Although this means that in most
cases, we terminate before the 10-minute timeout is reached (cf.\
\secref{sec:auto-experiments}), we believe this is a sensible decision, as we
observe that \gpt is unable to visually refine its outputs successfully in any
case. We hypothesize that this limitation is due to general-purpose chat models
requiring too much explicit context for this task. These models receive the
entire previously generated code as input, along with two input images and a
complex textual prompt, which may be too challenging for them to process
effectively. Preliminary experiments with more elaborate prompts did not seem
to mitigate the subpar performance, likely due to this reason.

\section{Annotator Demographics}\label{sec:demographics}
Our annotator team consists of eleven experts with extensive research
experience in science and technology. The team comprises one male faculty
member, two female PhD students, seven male PhD students, and one male research
assistant from another institution. We chose to work exclusively with expert
annotators based on the findings of \citet{belouadi2024automatikz}, which
demonstrated that crowd annotators often lack the necessary research background
to produce reliable annotations.

\section{Examples}\label{sec:examples}
To provide a better understanding of our work, we present a variety of examples
in this section.\ \tabref{tab:sketchfig} displays exemplary figures and
real sketches from \sketchfig, while \tabref{tab:datikz} shows figures and
synthetic sketches from \datikz[v2]. Additionally,
\tabtabref{tab:human-outputs}{tab:auto-outputs} present sample outputs
generated by our systems during our human and automatic evaluations.\
\figref{fig:code} provides a closer look at generated code.

When comparing the real sketches in \tabref{tab:sketchfig} to their
corresponding reference figures, it becomes evident that the sketches often
contain less detail. For instance, sketches may lack colors or grids and
feature less precise lines. Moreover, the handwritten nature of the
sketches can sometimes make the text within them harder to read. These
characteristics are also present in the synthetic sketches shown in
\tabref{tab:datikz}. However, the problem of illegible text is more pronounced
in these sketches, as generating readable text remains a common challenge for
image generation models~\citep{borji2023deepfakes}. While the text may still
retain its meaning in a hidden way~\citep{daras2022discovering}, this could
lead to hallucinated text in the generated \tikzname programs. Nonetheless, we
believe that this aspect can still be advantageous for end users, as it enables
them to quickly add scribbles to indicate the desired text placement. By doing
so, \projectname can generate code for the overall structure and layout,
allowing users to easily modify and replace the text afterward.

The randomly selected generated figures from our human and automatic
evaluations~(cf.\ \secref{sec:human-eval} \& \secref{sec:auto-experiments})
shown in \tabtabref{tab:human-outputs}{tab:auto-outputs} corroborate our
quantitative findings.\@ \projectname[DS][7b]~(TI) demonstrates the best
overall performance and shows the least amount of fidelity errors, confirming
the effectiveness of our \ssim*-based \mcts refinement algorithm. However, we
still observe some inconsistencies, such as in layout and axes labeling,
although to a lesser extent compared to \projectname[DS][7b]~(OI) and \gpt. We
attribute the prevalence of this problem partly to our focus on perceptual
similarity rather than, e.g., pixel-level similarity, which allows the models
greater flexibility in interpreting the general semantics of the input figures
and sketches. While optimizing pixel-level similarity could be an alternative
approach, we argue that perceptual similarity can serve as a more meaningful
measure, especially when considering sketches. We believe that real users who
provide rough sketches of unfinished ideas will find the generated outputs that
interpret and refine their concepts to be inspirational. However, we
acknowledge the potential benefits of exploring more rigorous similarity
measures and plan to investigate this in future research.
Interestingly, \gpt occasionally generates outputs that may not be appropriate
in a scientific context, such as mistakenly embedding a smiley face in the
fourth example in \tabref{tab:human-outputs}. Instead of resolving such issues,
\gpt~(TI) further emphasizes these details, distancing the output from the
actual reference.

\figref{fig:code} provides a side-by-side comparison of the generated \tikzname
programs corresponding to the first row in \tabref{tab:human-outputs}.
\projectname[DS][7b] demonstrates its ability to utilize advanced abstractions
and control flow statements, generating code that is free of compile-time
errors in both OI and TI configurations. On the other hand, \gpt~(OI)
incorrectly uses an undefined arrow tip kind \verb|stealth'| in lines 9 and 10,
resulting in recoverable compile-time errors.\ \gpt~(TI) contains the same
error in line 8 and introduces additional errors in lines 16 and 26, where the
\verb|*| symbol would have to be removed from the loop lists for successful
expression evaluation.

\begin{table*}
  \setlength\q{\dimexpr .5\textwidth -2\tabcolsep}
  \begin{tabular}{*{2}{w{c}{\q}}}
    \toprule
    \thead{Reference Figures} & \thead{Real Sketches}\\
    \midrule
    \includegraphics[height=4cm]{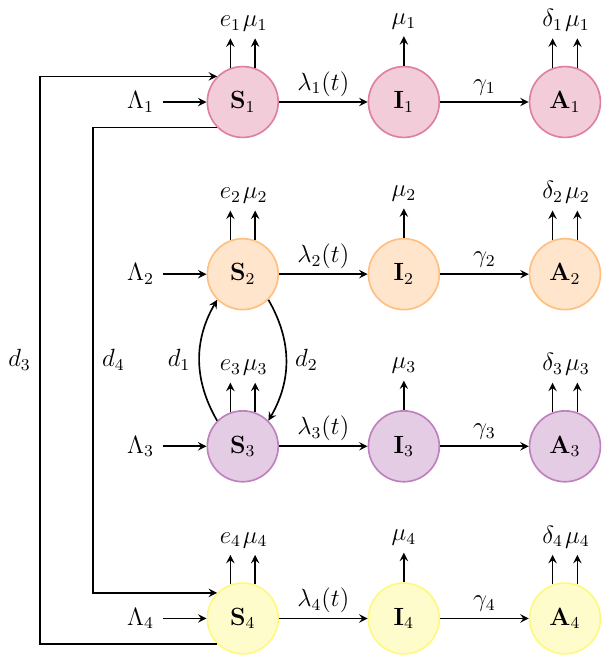} & \includegraphics[height=4cm]{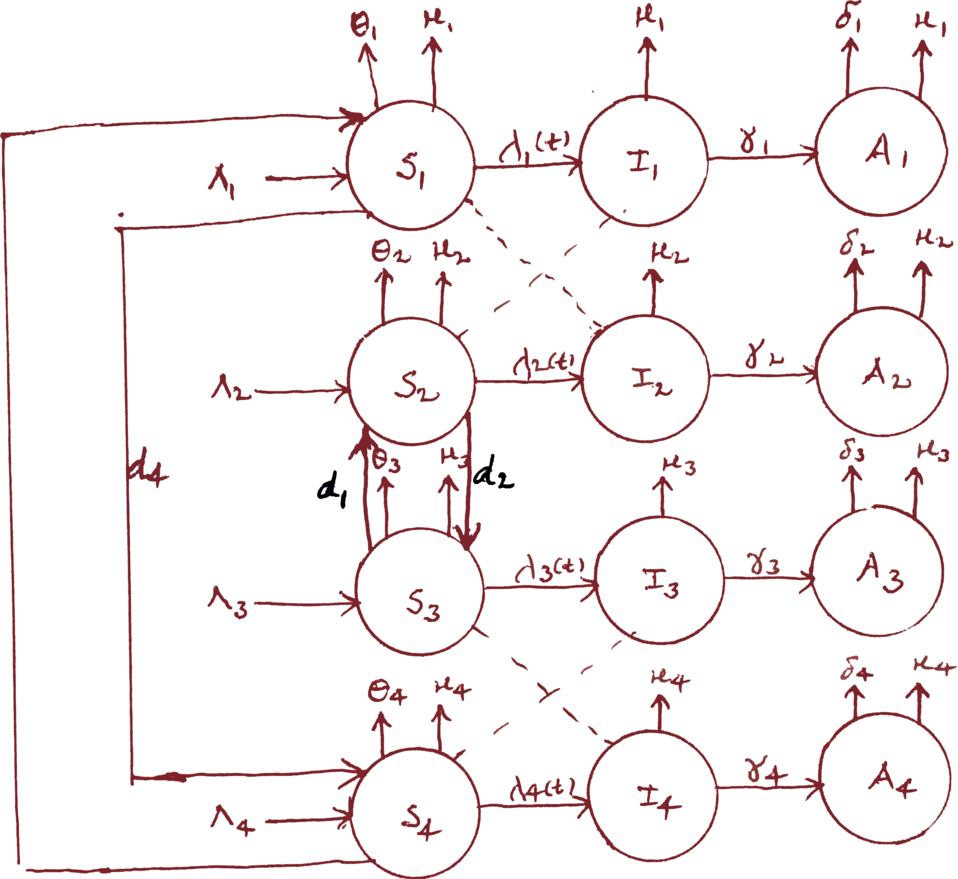} \\
    \midrule
    \includegraphics[height=4cm]{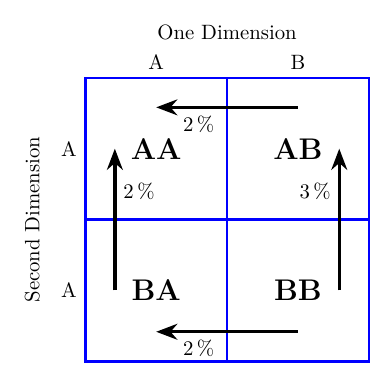} & \includegraphics[height=4cm]{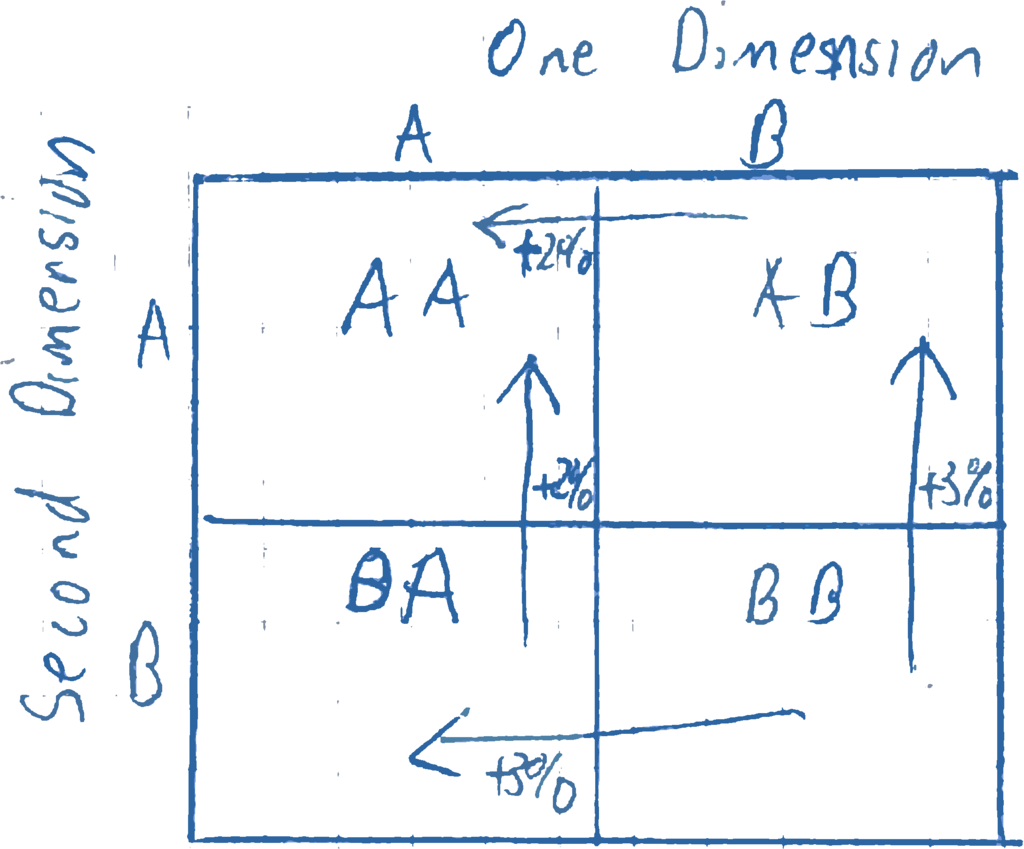} \\
    \midrule
    \includegraphics[height=4cm]{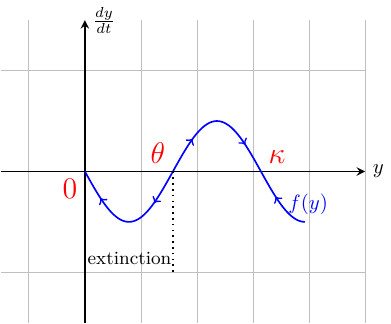} & \includegraphics[height=4cm]{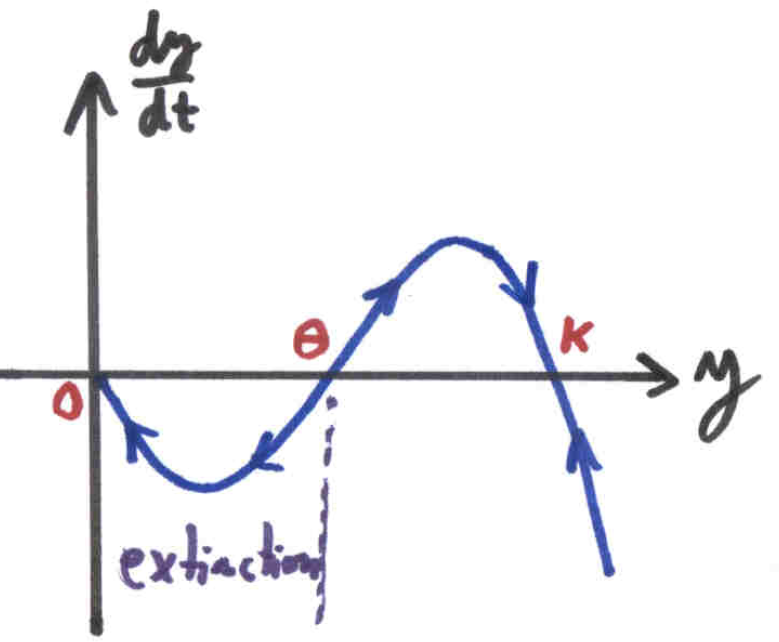} \\
    \midrule
    \includegraphics[height=4cm]{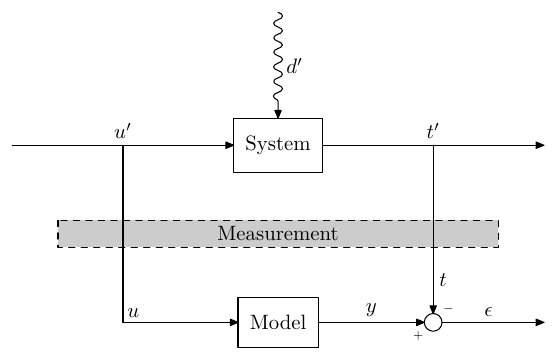} & \includegraphics[height=4cm]{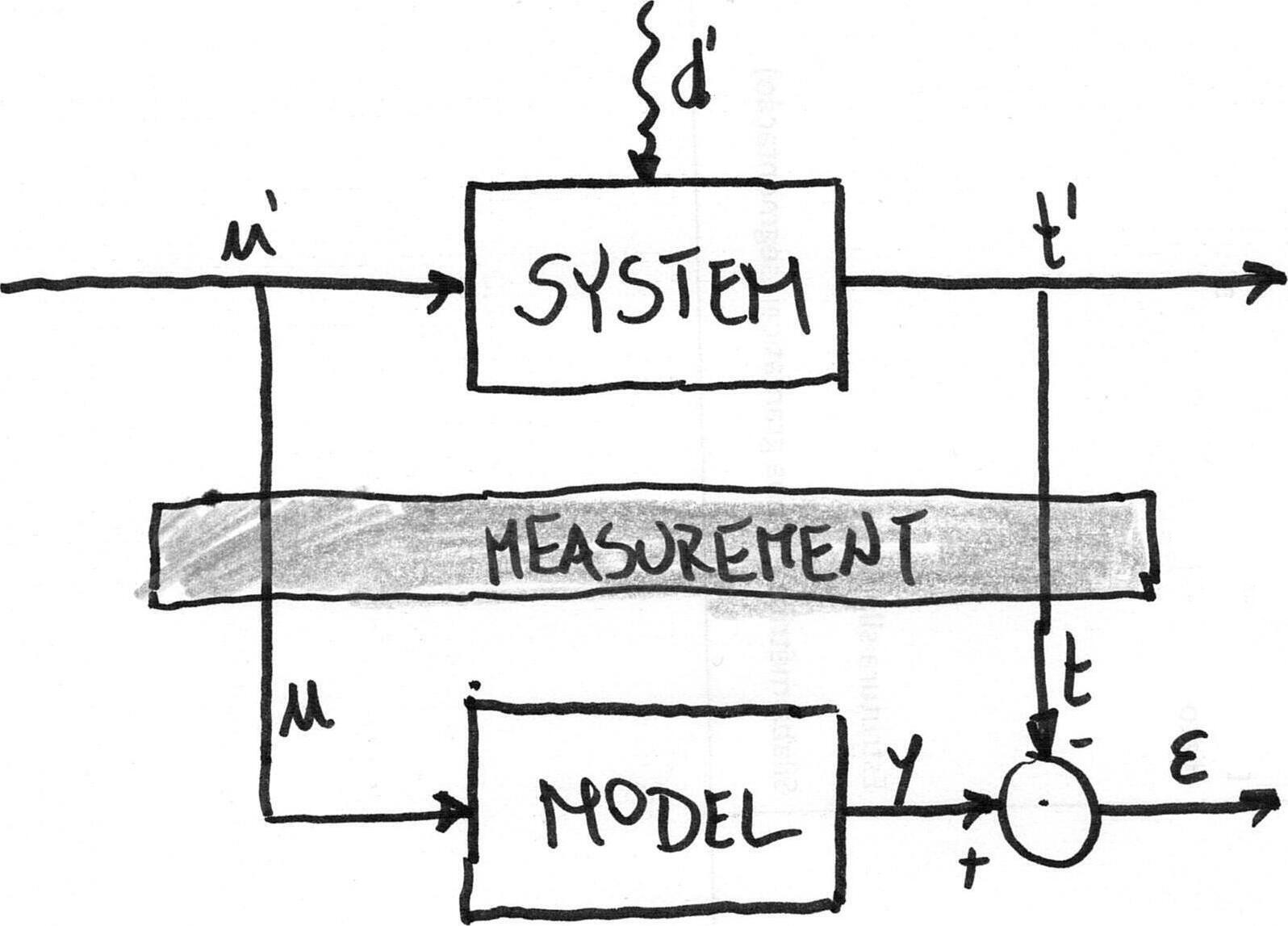} \\
    \bottomrule
  \end{tabular}
    \caption{Representative examples of reference figures paired with real
      sketches from the \sketchfig dataset.}%
  \label{tab:sketchfig}
\end{table*}
\begin{table*}
  \setlength\q{\dimexpr .5\textwidth -2\tabcolsep}
  \begin{tabular}{*{2}{w{c}{\q}}}
    \toprule
    \thead{Reference Figures} & \thead{Synthetic Sketches}\\
    \midrule
    \includegraphics[height=4cm]{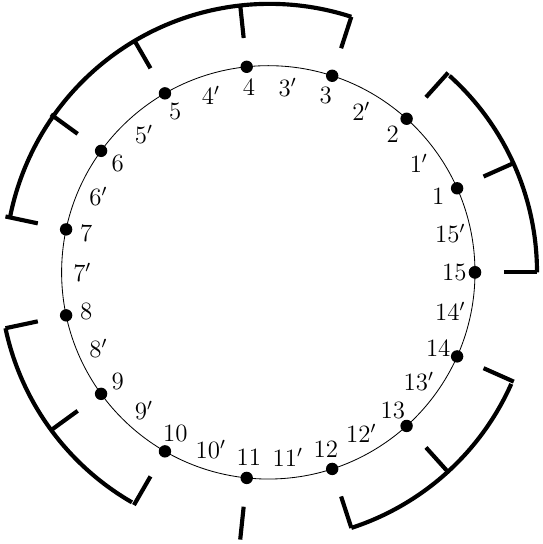} & \includegraphics[height=4cm]{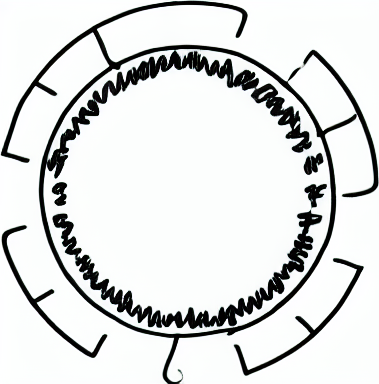}\\
    \midrule
    \includegraphics[height=4cm]{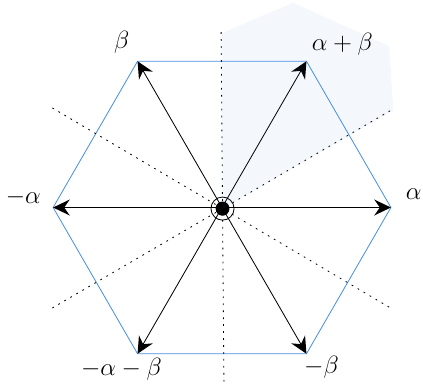} & \includegraphics[height=4cm]{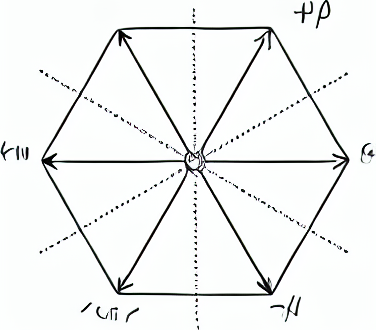}\\
    \midrule
    \includegraphics[height=4cm]{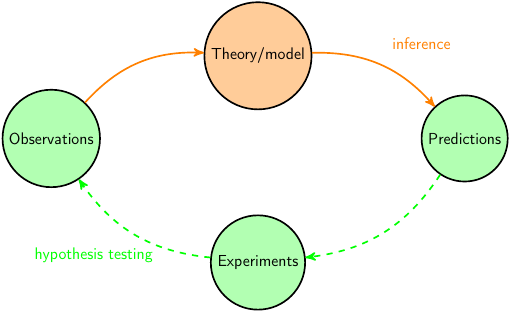} & \includegraphics[height=4cm]{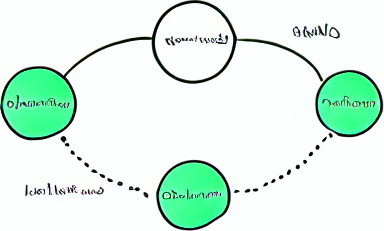}\\
    \midrule
    \begin{minipage}[b][4cm][c]{\q}%
      \includegraphics[width=\textwidth]{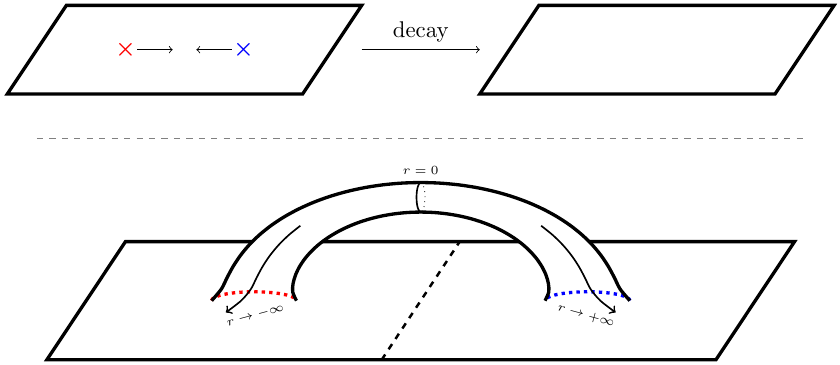}%
    \end{minipage} &
    \begin{minipage}[b][4cm][c]{\q}%
      \includegraphics[width=\textwidth]{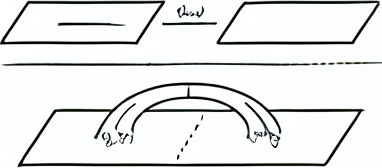}%
    \end{minipage}\\
    \bottomrule
  \end{tabular}
  \caption{Illustrative examples of reference figures and corresponding
    synthetic sketches from the subset of the \datikz[v2] dataset that is
    licensed for redistribution.}%
  \label{tab:datikz}
\end{table*}
\begin{table*}
  \scriptsize
  \pgfmathsetlength{\q}{\textwidth/5 -2\tabcolsep}
  \newcommand{\includerow}[2][png]{%
    \includegraphics[height=2cm,width=\q,keepaspectratio]{evaluation/human/#2/human_#1.#1} &
    \includegraphics[height=2cm,width=\q,keepaspectratio]{evaluation/human/#2/gpt-4-textual_pdf.pdf} &
    \includegraphics[height=2cm,width=\q,keepaspectratio]{evaluation/human/#2/gpt-4-visual_pdf.pdf} &
    \includegraphics[height=2cm,width=\q,keepaspectratio]{evaluation/human/#2/detikzify-ds-7b-fast_pdf.pdf} &
    \begin{minipage}[b][2cm][c]{\q}
      \includegraphics[height=2cm,width=\q,keepaspectratio]{evaluation/human/#2/detikzify-ds-7b-timeout_pdf.pdf}%
    \end{minipage}\\
  }
  \begin{tabular}{*{5}{>{\centering\arraybackslash}m{\q}}}
    \toprule
    \thead{Input} & \thead{\gpt (OI)} & \thead{\gpt (TI)} & \thead{\projectname[DS][7b] (OI)} & \thead{\projectname[DS][7b] (TI)}\\
    \midrule
    \includerow[pdf]{1}
    \midrule
    \includerow{2}
    \midrule
    \includerow[pdf]{3}
    \midrule
    \includerow{4}
    \midrule
    \includerow[pdf]{5}
    \midrule
    \includerow{6}
    \midrule
    \includerow[pdf]{7}
    \midrule
    \includerow{8}
    \bottomrule
  \end{tabular}
  \caption{Alternating rows of randomly selected reference figures and real
  sketches~(first column) alongside corresponding scientific figures generated
  by \gpt and \projectname[DS][7b] in output-driven~(OI) and
  time-budgeted~(TI) configurations~(columns 2--4), taken from our human
  evaluation campaign~(cf.\ \secref{sec:human-eval}).}%
  \label{tab:human-outputs}
\end{table*}
\begin{table*}
  \scriptsize
  \pgfmathsetlength{\q}{\textwidth/5 -2\tabcolsep}
  \newcommand{\includerow}[2][png]{%
    \includegraphics[height=2cm,width=\q,keepaspectratio]{evaluation/auto/#2/input_#1.#1} &
    \includegraphics[height=2cm,width=\q,keepaspectratio]{evaluation/auto/#2/claude-3-fast_pdf.pdf} &
    \includegraphics[height=2cm,width=\q,keepaspectratio]{evaluation/auto/#2/gpt-4-fast_pdf.pdf} &
    \includegraphics[height=2cm,width=\q,keepaspectratio]{evaluation/auto/#2/detikzify-ds-7b-fast_pdf.pdf} &
    \begin{minipage}[b][2cm][c]{\q}%
      \includegraphics[height=2cm,width=\q,keepaspectratio]{evaluation/auto/#2/detikzify-ds-7b-timeout_pdf.pdf}%
    \end{minipage}\\
  }
  \begin{tabular}{*{5}{>{\centering\arraybackslash}m{\q}}}
    \toprule
    \thead{Input} & \thead{\claude (OI)} & \thead{\gpt (OI)} & \thead{\projectname[DS][7b] (OI)} & \thead{\projectname[DS][7b] (TI)}\\
    \midrule
    \includerow[pdf]{1}
    \midrule
    \includerow{2}
    \midrule
    \includerow[pdf]{3}
    \midrule
    \includerow{4}
    \midrule
    \includerow[pdf]{5}
    \midrule
    \includerow{6}
    \midrule
    \includerow[pdf]{7}
    \midrule
    \includerow{8}
    \bottomrule
  \end{tabular}
  \caption{Alternating rows of randomly selected reference figures and
  synthetic sketches~(first column) alongside corresponding scientific figures
  generated by \claude~(OI), \gpt~(OI), and \projectname[DS][7b]~(OI \& TI) in
  columns 2--4, taken from our automatic evaluation~(cf.\
  \secref{sec:auto-experiments}).}%
  \label{tab:auto-outputs}
\end{table*}
\begin{figure}
  \newtcbinputlisting{\examplelisting}[3][]{
    enhanced,%
    size=small,%
    fontupper=\tiny,%
    fonttitle=\scriptsize,%
    left=0pt,%
    right=0pt,%
    top=0pt,%
    bottom=0.3mm-\tcboxedtitleheight,%
    listing only,%
    minted language=latex,%
    minted options={%
      linenos=true,%
      firstnumber=1,%
      numbersep=2mm,%
      breaklines=true,%
      breaksymbol={},%
      breakaftersymbolpre={},%
      breakafter={,},%
      breakindentnchars=4,%
      firstline=2,%
      highlightcolor=TUDa-6a,%
      #1,
    },%
    attach boxed title to bottom right={%
      xshift=-0.3mm+0.1pt,
      yshift*=\tcboxedtitleheight+0.3mm-0.1pt%
    },%
    boxed title style={sharp corners=uphill,%
      size=small,%
      no borderline,%
      rightrule=0.1pt,%
      bottomrule=0.1pt%
    },%
    title={#2},%
    listing file={graphics/examples/evaluation/human/1/#3},%
  }%
  \centering
  \begin{tcbraster}[raster columns=2,raster equal height=rows]
    \examplelisting{\projectname[DS][7b] (OI)}{detikzify-ds-7b-fast.tex}
    \examplelisting[numbers=right]{\projectname[DS][7b] (TI)}{detikzify-ds-7b-timeout.tex}
    \examplelisting[highlightlines={9,10}]{\gpt (OI)}{gpt-4-textual.tex}
    \examplelisting[numbers=right,highlightlines={8,16,26}]{\gpt (TI)}{gpt-4-visual.tex}
  \end{tcbraster}
  \caption{\tikzname programs generated by \projectname[DS][7b]~(top) and
    \gpt~(bottom) corresponding to the figures in the first row of
    \tabref{tab:human-outputs}. Lines with compile-time errors are highlighted
    in yellow.}%
  \label{fig:code}
\end{figure}